\documentclass[preprint,12pt,3p]{elsarticle}
\usepackage{lineno}
\usepackage{float}
\usepackage{graphicx}
\usepackage{multirow}
\usepackage{array}
\modulolinenumbers[5]
\journal{}

\usepackage{colortbl}
\usepackage{verbatim}
\usepackage{amsmath,amsthm,amsfonts,amssymb}
\usepackage{algorithm}
\usepackage{array}
\usepackage{enumerate}
\usepackage{graphics}
\usepackage{natbib}
\usepackage{algpseudocode}
\usepackage{hyperref}
\usepackage{mathrsfs}
\usepackage{multicol}
\usepackage{multirow}
\usepackage{subcaption}
\newcolumntype{C}[1]{>{\centering\let\newline\\\arraybackslash\hspace{0pt}}m{#1}}
\def\BibTeX{{\rm B\kern-.05em{\sc i\kern-.025em b}\kern-.08em
    T\kern-.1667em\lower.7ex\hbox{E}\kern-.125emX}}
\begin{document}

\begin{frontmatter}

\title{Atmospheric Noise-Resilient Image Classification in a Real-World Scenario: Using Hybrid CNN and Pin-GTSVM}

\author{
Shlok Mehendale$^{a}$, Jajati Keshari Sahoo$^{b}$\footnote{Corresponding author}, Rajendra Kumar Roul$^{c}$}
\address{$^a$Department of Computer Science $\&$ Engineering, BITS Pilani K K Birla Goa Campus, Goa, India,  f20221426@goa.bits-pilani.ac.in.\\$^b$Department of Mathematics, BITS Pilani K K Birla Goa Campus, Goa, India, jksahoo@goa.bits-pilani.ac.in\\$^c$Department of Computer Science $\&$ Engineering, Thapar Institute of Engineering $\&$ Technology, Patiala, Punjab, India, raj.roul@thapar.edu
}

\begin{abstract}
Parking space occupation detection using deep learning frameworks has seen significant advancements over the past few years. While these approaches effectively detect partial obstructions and adapt to varying lighting conditions, their performance significantly diminishes when haze is present.  This paper proposes a novel hybrid model with a pre-trained feature extractor and a Pinball Generalized Twin Support Vector Machine (Pin-GTSVM) classifier, which removes the need for a dehazing system from the current State-of-The-Art hazy parking slot classification systems and is also insensitive to any atmospheric noise. The proposed system can seamlessly integrate with conventional smart parking infrastructures, leveraging a minimal number of cameras to monitor and manage hundreds of parking spaces efficiently. Its effectiveness has been evaluated against established parking space detection methods using the CNRPark Patches, PKLot, and a custom dataset specific to hazy parking scenarios. Furthermore, empirical results indicate a significant
improvement in accuracy on a hazy parking system, thus emphasizing efficient atmospheric noise handling. 
\end{abstract}


\end{frontmatter}

\section{Introduction}
In recent years, image classification has become crucial in many real-world applications, such as autonomous driving, surveillance, and environmental monitoring. In densely populated urban areas, studies have shown that nearly 30\% of vehicular traffic consists of drivers searching for parking spaces \cite{aftab2020reducing}. This persistent quest for parking places significant strain on parking management systems and leads to prolonged commute times as drivers struggle to locate available spots \cite{mangiaracina2017smart,liao2024multi}. Implementing effective parking classification methods is essential to mitigating these challenges. These methods provide valuable data on parking space availability, aiding in efficient parking lot operations. By offering real-time information about vacant spots, the system can help to minimize waiting times, enhance operational scalability, and reduce the driver's time for searching the free parking slot \cite{thakur2024deep}. For many, particularly inexperienced drivers, parking can be a daunting task. This difficulty is heightened under conditions of poor visibility, such as during foggy weather, where assessing the dimensions of parking spaces and identifying nearby obstacles becomes significantly more challenging\footnote{https://www.sciencedirect.com/science/article/pii/S0957417423006498}.  \\

Advancements in Deep Learning, particularly with the development of pre-trained models such as ResNet-50, GoogleNet, and AlexNet, have significantly enhanced the performance of image classification tasks. However, real-world scenarios often present challenging conditions like atmospheric noise, which can greatly degrade the quality of images captured by sensors and cameras. This noise, which includes fog, rain, snow, and dust, introduces significant variability and corruption into the image data, making it difficult for conventional image classification models to maintain high accuracy and reliability. Vision-based detection of parking slot occupancy is widely regarded as a highly scalable and cost-efficient solution, efficiently utilizing less number of cameras to oversee and manage hundreds of parking spaces \cite{b9, b10, b11, nc20}. However, the performance of such techniques is adversely affected by factors such as obstructions (e.g., trees blocking parking slots) and changing weather conditions. To address these challenges, studies like \cite{b10, b11} have leveraged the robustness of Deep Neural Networks (DNN) to improve detection reliability. Shoup et al. \cite{b2} and Giuffre et al.\cite{c101} reported that vehicles searching for parking spaces account for a significant portion of urban traffic, contributing approximately 25\% to 40\% of the total volume. Similarly, Lin et al. \cite{b3} point out that distracted drivers seeking parking spaces significantly increase the likelihood of accidents. These findings underline the importance of implementing smart parking systems to provide updates on real-time parking availability, reduce search times, and improve road safety. While promising, these method faces challenges with lighting variations and struggle to detect vehicles in the presence of obstructions, particularly in hazy conditions.  Due to fog, visibility becomes very low, with a high chance of misclassifying parking spaces. 

\subsection{Motivation}

The primary motivation behind this work is the need to develop a robust image classification system that can operate effectively under adverse environmental conditions.  While traditional models are powerful, they often struggle to maintain performance when faced with atmospheric noise due to their sensitivity to image quality and lack of specialized mechanisms to handle such distortions. Their performance drop in noisy environments can have serious implications, especially in critical applications like autonomous vehicles, where misclassifications can lead to safety hazards.

To address these challenges, this study proposes a novel hybrid approach that combines pre-trained models of CNN such as ResNet-50 \cite{resnet50}, GoogleNet \cite{GNet}, and AlexNet \cite{AlexNet} with Pinball Generalized Twin Support Vector Machines (Pin-GTSVM) \cite{pingtsvm} for parking slot classification under hazy conditions. The Pin-GTSVM is an advanced machine learning model designed for classification tasks, particularly in handling imbalanced data. It extends the Twin Support Vector Machine (TSVM) by incorporating the pinball loss function, which improves robustness against outliers and noise. Pin-GTSVM effectively enhances classification performance by focusing on quantile regression principles, making it well-suited for applications where asymmetric error handling is critical. The proposed hybrid model leverages the feature extraction capabilities of these pre-trained models and the robustness of Pin-GTSVM to atmospheric noise, aiming to improve classification accuracy in noisy conditions. 

A significant application of this research is demonstrated in hazy parking slot detection, which is a critical component of autonomous parking systems.  Traditional methods often rely on image dehazing techniques to preprocess images before applying classification algorithms. However, these dehazing processes can be computationally expensive and may not always produce satisfactory results.  By integrating the strengths of pre-trained models and Pin-GTSVM, the proposed hybrid model can directly classify parking slots in hazy conditions \textit{eliminating the need for a separate dehazing step or a Dehazer} in contrast with the existing work \cite{refpaper}. This simplifies the processing pipeline and enhances the system's efficiency and reliability in real-world scenarios.

Previous studies have explored various methods to enhance the resilience of image classification models to noise. For instance, some approaches have incorporated noise reduction techniques \cite{DnCNN} or augmented training datasets with artificially generated noisy images to improve model robustness\cite{hendrycks}. However, these methods often involve significant preprocessing overhead or do not generalize well to different types of atmospheric noise \cite{batchrenorm}. The hybrid approach proposed in this study seeks to overcome these limitations by integrating the strengths of the pre-trained models with Pin-GTSVM in a complementary manner, thereby achieving higher classification accuracy with reduced sensitivity to noise.

\subsection{Contribution}

The key contributions of this research are as follows:
\begin{enumerate}
    \item[i.] Hybrid Model Development: Designing and implementing a novel hybrid model that combines pre-trained models (ResNet-50, GoogleNet, and AlexNet) with Pin-GTSVM to leverage their respective strengths.
    \item[ii.] Noise Resilience: Evaluating the noise resilience of the proposed model in various atmospheric conditions through extensive experimentation.
    \item[iii.] Performance Benchmarking: Comparing the performance of the hybrid model with state-of-the-art image classification methods under noisy and general conditions to demonstrate its effectiveness.
    \item[iv.] Application in Hazy Parking Slot Detection: Demonstrating the practical application of the hybrid model in detecting parking slots in hazy conditions without requiring dehazing, thus simplifying the processing pipeline and improving system efficiency.
\end{enumerate}

The paper is outlined as follows: Section 2 delineates the literature review, whereas Section 3 addresses the preliminaries. Section 4 outlines the methodology of the suggested strategy. Section 5 discusses the findings of the experimental research. Section 6 concludes the report with prospective enhancements.

\section{Related Work}
 \subsection{Studies on Pin-GTSVM}
 The General Twin Support Vector Machine with Pinball Loss was proposed by Tanveer et al. \cite{pingtsvm} as an alternative to the standard twin support vector machine with a hinge loss function. An extension of the same is Pin-RGTSVM \cite{RobustPingtsvm}, which implements the structural risk minimization principle, increasing the robustness of the model to noisy datasets, and the matrices that appear in the dual formation here are positive definite compared to the earlier semi-definite matrices. Further variants of the Pinball TWSVM include models like LPTWSVM \cite{Tanveer2022}, demonstrating its scalability and applications by eliminating the need to calculate inverse matrices in the dual problem (which are very computationally demanding and also may not be possible due to matrix singularity); a Pinball Loss Twin Support Vector Clustering algorithm \cite{PinTSVC} where the pinball loss function replaces the hinge loss function and a Pinball Twin Bounded Support Vector Clustering algorithm \cite{PinTBSVC}, where unlike PinTSVC it has nonsingular matrices and builds on bounded twin support vector machines. Some sparsity-addressing versions include the Sparse Twin Support Vector clustering \cite{SparseTSVC} that involves the $\epsilon$-insensitive pinball loss function to formulate a sparse solution and the Improved SPTWSVM \cite{ImSpPinTWSVM} that simply inculcates the structural risk minimization (SRM) principle.
\subsection{Studies on CNN and  Pre-trained models for parking slot detection} \label{2.2}
 Prova et al. \cite{prova2022real} proposed a model using CNN for real-time indoor and outdoor environment classification. The study revealed that the model achieved higher accuracy in outdoor environments than indoor ones. Additionally, the model demonstrated strong generalizability, as it was trained in indoor environments and successfully tested in outdoor settings. Dhope et al. \cite{dhope2021novel} introduced a novel two-phase parking space tracking system that combined Mask R-CNN and YOLOv3 for parking slot detection. Their approach was validated under three distinct weather conditions: sunny, rainy, and overcast by measuring the Minimum Detection Rate (MDR). Rafique et al. \cite{rafique2023optimized} explored a vehicle detection system using a pre-trained YOLOv5 model, moving beyond classifying parking spaces as occupied or vacant. This system achieved an impressive accuracy of 99.5\% when evaluated on the PKLot dataset (de Almeida et al. \cite{de2015pklot}).  An updated version of AlexNet’s architecture (Krizhevsky et al. \cite{c100}) was also analyzed for its ability to recognize objects under challenging conditions, such as lighting variations and the presence of obstacles. The study demonstrated that the modified architecture effectively handled these challenges, showcasing its robustness in dynamic scenarios. Amato et al. \cite{amato2017deep} employed a CNN-based model for the classification of parking space, and the proposed CNRPark-EXT dataset contains data from different viewpoints, occlusion, and shadow situations, which makes parking space detection a challenging task. The CNN model for real-time parking space classification was proposed by Nyambal et al. \cite{nyambal2017automated}. The model undergoes training using the LeNet architecture with Nesterov's Accelerated Gradient for one part and the AlexNet network with Stochastic Gradient Descent for another. The model attains 99\% accuracy on the validation set with both networks. Further studies on smart parking systems using CNN and R-CNN techniques can be found in \cite{diaz2020survey, barriga2019smart,trivedi2020different, min2021attentional}. 

 \par While the works mentioned in Section \ref{2.2} have primarily focused on non-hazy conditions for parking space classification, the task becomes notably more challenging in hazy conditions. This study proposes a hybrid approach that addresses these limitations by combining the strengths of pre-trained models with Pin-GTSVM, where the pre-trained model is used for feature extraction and pin-GTSVM for classification. This integration enhances classification accuracy while minimizing sensitivity to noise and completely eliminates the need for a separate dehazing process which is generally required during hazy conditions.
 
\section{Preliminaries and Background}
\begin{figure}[H] 
    \centering
    \includegraphics[width=50mm]{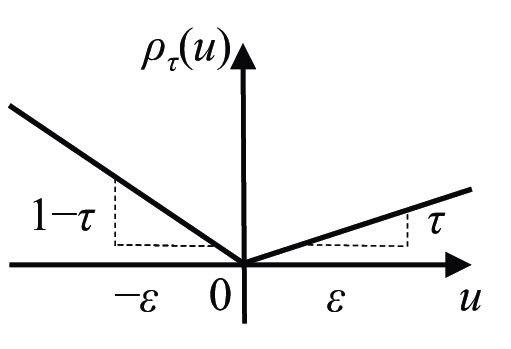} 
    \caption{Pinball Loss Function\cite{pin78}}
    \label{fig:1}
\end{figure}

This section presents the foundational concepts underlying the methodology adopted in the proposed research work, which explore the Pinball Loss function (shown in Figure \ref{fig:1}), the Twin Support Vector Machine (TWSVM) suggested by Jayadeva et al. \cite{twsvm}, and its variant, Pin-GTSVM, proposed by M. Tanveer et al. \cite{pingtsvm}. Additionally, we explored various pre-trained CNN models, including ResNet-50, AlexNet, and GoogleNet, which serve as key components of the proposed approach.

\subsection{Pinball Loss Function}

A loss function quantifies the difference between predictions and ground truth labels, guiding the optimization during model training. In the proposed research, we used the Pinball Loss function, which is employed along with the TWSVM as a part of Pin-GTSVM. The Pinball Loss penalizes deviations from specific quantiles $\tau$ of the target distribution, providing robustness against outliers and heteroscedasticity. The quantile levels considered here are 0.5, 0.8, and 1. Specifically, the Pinball Loss is formulated as shown in equation \ref{eq1}:

\begin{equation}
L_\tau(x, y, f(x)) = 
\begin{cases} 
-\tau (1 - y f(x)), & \text{if } 1 - y f(x) < 0, \\
(1 - \tau)(1 - y f(x)), & \text{if } 1 - y f(x) \ge 0.
\end{cases}
 \label{eq1}
\end{equation}
where \(0 < \tau < 1\) and
\begin{itemize}
    \item \( y \) represents the true label.
    \item \( f(x) \) denotes the predicted value.
   
\end{itemize}

For instance, setting \( \tau \) to 0.5 converts the function into Mean Absolute Error (MAE), targeting the median. The Pinball Loss function is instrumental in developing robust prediction intervals, making it highly applicable in fields requiring precise quantile estimations. This function ensures that the model remains sensitive to the specific quantiles of interest, thus enhancing the reliability of predictions.

\subsection{Twin Support Vector Machine (TWSVM)}
The TWSVM constructs two non-parallel proximal hyperplanes, where the data points of one class influence the constraints for the hyperplane associated with the opposite class. By decomposing the optimization task into two smaller Quadratic Programming Problems (QPPs) rather than a single larger one, TWSVM achieves a notable computational efficiency, typically performing up to four times faster than the standard Support Vector Machine (SVM) \cite{twsvm}.

For a binary classification problem with two classes labeled as +1 and -1, let the data points corresponding to class +1 and class -1 be represented in $\mathbb{R}^n$ by matrices $A$ and $B$, respectively. Here, $A$ contains $l_1$ points, while $B$ consists of $l_2$ points. Nonlinear TWSVM seeks to construct two kernel-induced surfaces, defined mathematically as follows:

\[
K(x^T, D^T) \mathbf{u}^{(+)} + b^{(+)} = 0 \quad \text{and} \quad K(x^T, D^T) \mathbf{u}^{(-)} + b^{(-)} = 0,
\]
where \( D = [A; B] \); \(\mathbf{u}^{+}, \mathbf{u}^{-} \in \mathbb{R}^n \) and \( K \) is an arbitrary kernel function.

The nonlinear TWSVM formulation can be expressed as follows:

\[
\min_{\mathbf{u}^{+}, b^{+}, \boldsymbol{\xi}_1} \frac{1}{2} \| K(A, D^T) \mathbf{u}^{+} + e_1 b^{+} \|^2 + c_1 e_2^T \boldsymbol{\xi}_1
\]
subject to
\[
- K(B, D^T) \mathbf{u}^{+} + e_2 b^{+} + \boldsymbol{\xi}_1 \geq e_2, \quad \boldsymbol{\xi}_1 \geq 0,
\]
and
\[
\min_{\mathbf{u}^{-}, b^{-}, \boldsymbol{\xi}_2} \frac{1}{2} \| K(B, D^T) \mathbf{u}^{-} + e_2 b^{-} \|^2 + c_2 e_1^T \boldsymbol{\xi}_2
\]
subject to
\[
K(A, D^T) \mathbf{u}^{-} + e_1 b^{-} + \boldsymbol{\xi}_2 \geq e_1, \quad \boldsymbol{\xi}_2 \geq 0,
\]
where \( c_1, c_2 \) are positive parameters, \( e_1, e_2 \) are vectors of ones of appropriate dimensions, and \( \boldsymbol{\xi}_1, \boldsymbol{\xi}_2 \) are slack variables.

TWSVMs are an extension of traditional Support Vector Machines (SVMs) designed for binary classification specifically. Unlike SVMs, which construct a single hyperplane, TWSVMs generate two non-parallel hyperplanes, each closer to one of the two classes. This approach solves two smaller QPPs instead of a single large one, leading to faster training times. The optimization problems for TWSVMs are formulated as:

\begin{align}
    \min_{\mathbf{w}_1, b_1} & \quad \frac{1}{2} \|\mathbf{X}_1 \mathbf{w}_1 + \mathbf{e}_1 b_1\|^2 + c_1 \mathbf{e}_2^\top (\mathbf{X}_2 \mathbf{w}_1 + \mathbf{e}_2 b_1 + \mathbf{e}_2) \\
    \text{s.t.} & \quad \mathbf{X}_2 \mathbf{w}_1 + \mathbf{e}_2 b_1 \geq \mathbf{e}_2
\end{align}

and

\begin{align}
    \min_{\mathbf{w}_2, b_2} & \quad \frac{1}{2} \|\mathbf{X}_2 \mathbf{w}_2 + \mathbf{e}_2 b_2\|^2 + c_2 \mathbf{e}_1^\top (\mathbf{X}_1 \mathbf{w}_2 + \mathbf{e}_1 b_2 + \mathbf{e}_1) \\
    \text{s.t.} & \quad \mathbf{X}_1 \mathbf{w}_2 + \mathbf{e}_1 b_2 \leq -\mathbf{e}_1
\end{align}

Here, \( \mathbf{X}_1 \) and \( \mathbf{X}_2 \) are the data matrices of the two classes, \( \mathbf{w}_1 \) and \( \mathbf{w}_2 \) are the weight vectors, \( b_1 \) and \( b_2 \) are the biases, \( \mathbf{e}_1 \) and \( \mathbf{e}_2 \) are vectors of ones, and \( c_1 \) and \( c_2 \) are regularization parameters.

\subsection{ \textit{Generalized Twin Support Vector Machine (Pin-GTSVM)}}
Pin-GTSVM extends the TWSVM framework by incorporating the Pinball loss function, enhancing robustness in classification tasks. The Pinball loss provides asymmetric penalties for overestimations and underestimations, controlled by the quantile parameter \( \tau \). Pin-GTSVM also solves two QPPs, each aiming to minimize the Pinball loss:

\begin{align}
    \min_{\mathbf{w}_1, b_1, \xi} & \quad \frac{1}{2} \|\mathbf{X}_1 \mathbf{w}_1 + \mathbf{e}_1 b_1\|^2 + c_1 \sum_{i=1}^{m_2} \left[ \tau \xi_i + (1 - \tau) \max(0, \xi_i) \right] \\
    \text{s.t.} & \quad \mathbf{X}_2 \mathbf{w}_1 + \mathbf{e}_2 b_1 \geq \mathbf{e}_2 - \xi, \quad \xi \geq 0
\end{align}

and

\begin{align}
    \min_{\mathbf{w}_2, b_2, \eta} & \quad \frac{1}{2} \|\mathbf{X}_2 \mathbf{w}_2 + \mathbf{e}_2 b_2\|^2 + c_2 \sum_{j=1}^{m_1} \left[ \tau \eta_j + (1 - \tau) \max(0, \eta_j) \right] \\
    \text{s.t.} & \quad \mathbf{X}_1 \mathbf{w}_2 + \mathbf{e}_1 b_2 \leq -\mathbf{e}_1 + \eta, \quad \eta \geq 0
\end{align}

Here, \( \mathbf{X}_1 \) and \( \mathbf{X}_2 \) are the data matrices of the two classes, \( \mathbf{w}_1 \) and \( \mathbf{w}_2 \) are the weight vectors, \( b_1 \) and \( b_2 \) are the biases, \( \mathbf{e}_1 \) and \( \mathbf{e}_2 \) are vectors of ones, \( \xi \) and \( \eta \) are slack variables, and \( c_1 \) and \( c_2 \) are regularization parameters. Pin-GTSVM leverages the robustness of Pinball loss, making it effective for tasks requiring precise quantile estimations in noisy environments.

\subsection{Resnet-50}
\begin{figure}[H] 
    \centering
    \includegraphics[width=\linewidth]{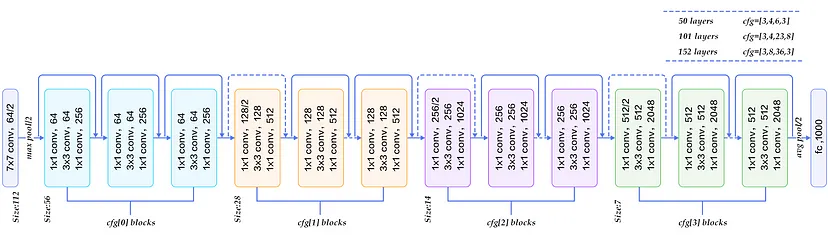} 
    \caption{Res-Net50 Architecture}
    \label{fig:2}
\end{figure}

He et al.\cite{resnet50}, who introduced ResNet-50 (Figure \ref{fig:2}) in the year 2015, is a deep CNN architecture. ResNet's primary innovation lies in its use of residual learning, effectively tackling the degradation problem commonly encountered in deep networks. This is accomplished through residual blocks, which enable more efficient gradient flow throughout the network by learning residual functions relative to the inputs of each layer.

ResNet-50 is a deep CNN architecture consisting of 50 layers, organized to facilitate efficient learning and gradient flow through its use of residual connections. The structure includes:

\begin{itemize}
    \item An initial convolutional layer with a 7$\times$7 filter and 64 output channels.
    \item A max-pooling layer.
    \item Contains 3 residual blocks, and each block has a bottleneck design: 1 $\times$ 1, 3 $\times$ 3, and 1 $\times$ 1 convolutional layers.
    \item The Global Average Pooling (GAP) layer is commonly used to reduce spatial dimensions and generate a fixed-length output regardless of the input size. 
    \item A fully connected layer with a softmax activation is commonly used for classification tasks.
\end{itemize}

The architecture can be summarized mathematically as:

\[
\mathbf{y} = \mathcal{F}(\mathbf{x}, \{ \mathbf{W}_i \}) + \mathbf{x}
\]

where \( \mathbf{x} \) is the input, \( \mathbf{y} \) is the output of the residual block, and \( \mathcal{F}(\mathbf{x}, \{ \mathbf{W}_i \}) \) represents the residual mapping to be learned. ResNet-50 has become a widely used backbone in many state-of-the-art image recognition models and it still performs the best when integrated with Pin-GTSVM.\newline

\subsection{AlexNet} 
\begin{figure}[H]
    \centering
    \includegraphics[width=\linewidth]{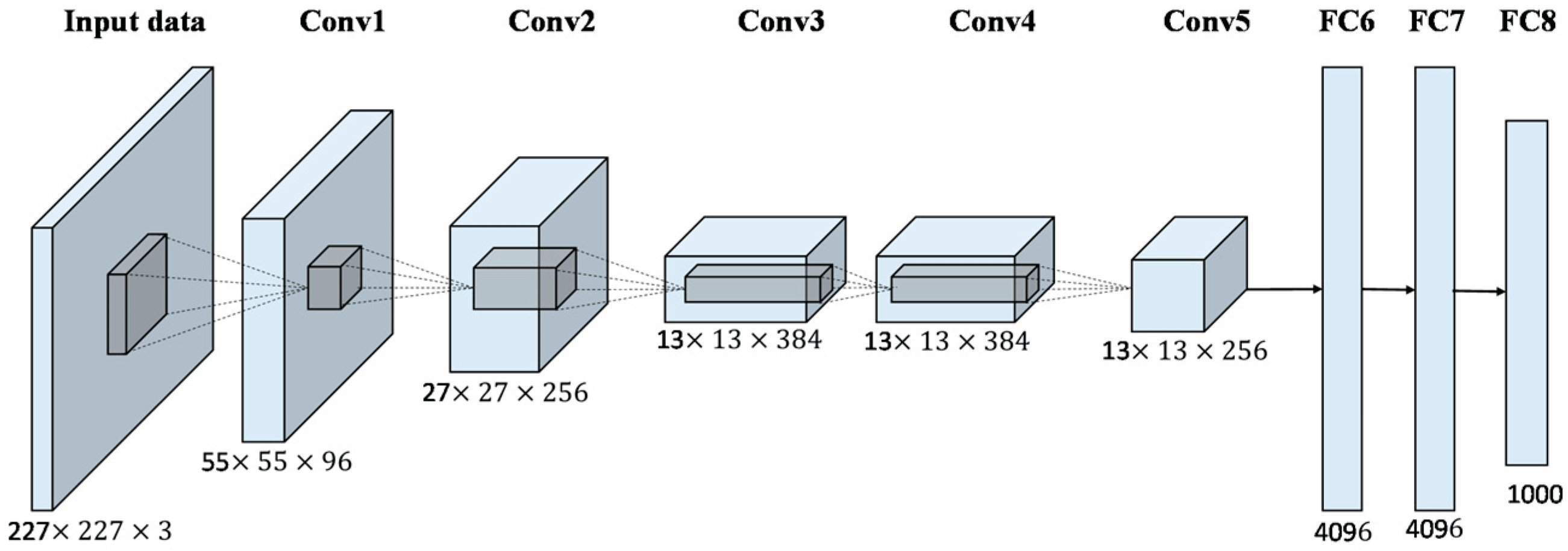}
    \caption{Overview of the AlexNet Architecture \cite{AlexNetimg}}
    \label{fig:alexnet_arch}
\end{figure}

AlexNet, introduced by Krizhevsky et al.\cite{AlexNet} in 2012, marked a significant breakthrough in the field of Computer Vision. This innovative CNN architecture, depicted in Figure \ref{fig:alexnet_arch}, consists of eight sequential layers: five convolutional layers dedicated to feature extraction, followed by three fully connected layers for classification.

The critical components of the AlexNet architecture are detailed below:

\begin{itemize}
    \item \textbf{Convolutional Layers:} The initial two convolutional layers utilize relatively large filter sizes (11$\times$11 and 5$\times$5), while the subsequent three layers adopt smaller 3$\times$3 filters to capture fine-grained features.
    \item \textbf{Max-Pooling Layers:} These layers follow the first, second, and fifth convolutional layers to downsample spatial dimensions, reduce computational complexity, and help mitigate overfitting.
    \item \textbf{Rectified Linear Unit (ReLU):} Applied after each convolutional and fully connected layer, the ReLU activation function introduces non-linearity, enabling the network to learn intricate patterns effectively.
    \item \textbf{Local Response Normalization (LRN):} To improve the generalization of the model, LRN is applied after the first and second convolutional layers.
    \item \textbf{Dropout:} Dropout regularization is employed after the first two fully connected layers, randomly deactivating a portion of neurons during training to reduce the risk of overfitting.
    \item \textbf{Data Augmentation:} Techniques such as random cropping, horizontal flipping, and color jittering are incorporated to artificially expand the diversity of the training dataset and improve model robustness.
\end{itemize}

The AlexNet architecture can be mathematically expressed as:

\[
\mathbf{y} = \text{softmax}(\mathbf{W}_3 \cdot \text{ReLU}(\mathbf{W}_2 \cdot \text{ReLU}(\mathbf{W}_1 \cdot \mathbf{x} + \mathbf{b}_1) + \mathbf{b}_2) + \mathbf{b}_3),
\]

where \( \mathbf{x} \) represents the input image, \( \mathbf{W}_i \) and \( \mathbf{b}_i \) denote the weights and biases of the \(i\)-th layer, and \( \mathbf{y} \) corresponds to the predicted class probabilities.

AlexNet’s introduction marked a significant milestone in deep learning, showcasing its power for large-scale image classification tasks. Its innovative use of the ReLU activation function, dropout regularization, and data augmentation strategies set new benchmarks and established foundational practices widely adopted in modern neural network designs.

\subsection{GoogLeNet} 
\begin{figure}[H]
    \centering
    \includegraphics[width=120mm]{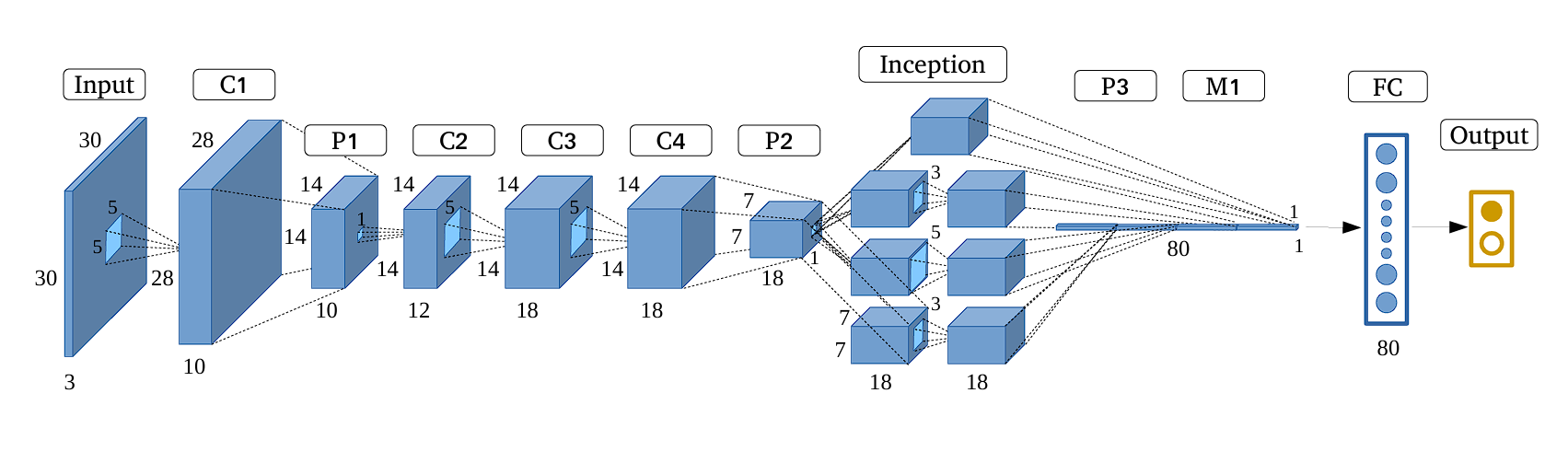} 
    \caption{GoogleNet Architecture}
    \label{fig:4}
\end{figure}

In 2014, Szegedy \textit{et al.} \cite{GNet} introduced GooLeNet, a deep CNN architecture. The hallmark of GoogLeNet is its Inception module (Figure \ref{fig:4}), which allows the network to capture multi-scale features effectively while keeping computational costs manageable.

The architecture can be summarized as a stack of multiple inception modules interleaved with max-pooling layers, followed by global average pooling and a final softmax layer for classification. 

GoogLeNet's innovative use of Inception modules made it possible to build a deep network (22 layers) with fewer parameters than similarly deep networks. This architecture set a new standard in CNN design and influenced many subsequent models in both academia and industry.

\subsection{Linear and Non-Linear Kernels}
In machine learning, kernel functions enable algorithms to operate in high-dimensional spaces without explicitly computing the data coordinates in that space. These functions accomplish this by calculating the inner products of the image representations for all data pairs within a feature space. We have utilized two widely used kernel types in the proposed approach, i.e., linear kernels and non-linear Gaussian (RBF) kernels.

\subsubsection{Linear Kernel}

The linear kernel is the simplest type of kernel function. It represents the standard inner product of two vectors in the input space, allowing linear algorithms to be applied directly to the original feature space. It is defined as:

\[
K(\mathbf{x}_i, \mathbf{x}_j) = \mathbf{x}_i^\top \mathbf{x}_j
\]

where \( \mathbf{x}_i \) and \( \mathbf{x}_j \) are input vectors. The linear kernel is particularly effective in cases where the data is linearly separable or when the dimensionality of the input space is very high.

\subsubsection{Gaussian (RBF) Kernel}

The Gaussian kernel \footnote{https://www.sciencedirect.com/topics/engineering/gaussian-kernel}, also known as the Radial Basis Function (RBF) kernel, is widely used for handling non-linear data. It transforms the input data into an infinite-dimensional feature space, enabling the algorithm to define intricate decision boundaries. The Gaussian kernel is defined as:

\[
K(\mathbf{x}_i, \mathbf{x}_j) = \exp\left(-\frac{\|\mathbf{x}_i - \mathbf{x}_j\|^2}{2\sigma^2}\right)
\]

where \( \|\mathbf{x}_i - \mathbf{x}_j\| \) is the Euclidean distance between the input vectors \( \mathbf{x}_i\) and \( \mathbf{x}_j \), and \( \sigma \) is a parameter that controls the width of the Gaussian function. The parameter \( \sigma \) determines the spread of the kernel and influences the decision boundary's flexibility. A small \( \sigma \) value leads to a decision boundary that tightly fits the data, while a large \( \sigma \) value results in a smoother decision boundary.

\section{Dataset Analysis}
The proposed system is tested on three widely used datasets: CNRPark patches, the PKLot dataset, and a custom Hazy Parking System dataset. The details are as follows:
\subsection{PKLoT Dataset: } 
\begin{figure}[hbt!]
    \centering
    \includegraphics[width=4in,height=1.5in]{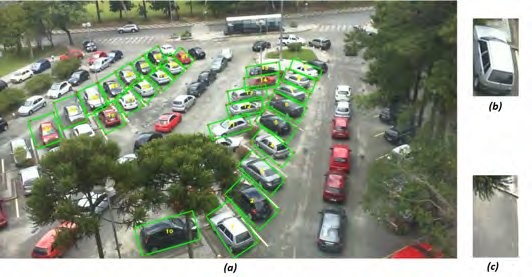}
    \caption{(a) Parking spot containing 28 delimited parking spaces, (b) occupied parking slot, and (c) empty parking slot. }
    \label{fig:PKlot}
\end{figure}

The PKLot dataset \cite{PKLot} is an upgraded version of the parking lot dataset initially developed by Almeida in 2013 \cite{almeida2013parking}. It includes images captured from parking facilities at PUCPR (Pontifical Catholic University of Paraná)\footnote{https://www.pucpr.br/international/} and UFPR (Federal University of Paraná)\footnote{https://internacional.ufpr.br/portal/about-ufpr/} in Curitiba, Brazil. To minimize visual obstructions caused by nearby vehicles, cameras were mounted on the rooftops of buildings. As shown in Figure \ref{fig:PKlot}, the dataset clearly distinguishes between occupied and vacant parking spots. 
  The PKLot dataset contains 12,417 images captured from two different parking lots on sunny, cloudy, and rainy days, accounting for variations of atmospheric noise.
 
\subsection{CNRPark Dataset:} 
\begin{figure}[h!!]
    \centering
    \includegraphics[width=50mm]{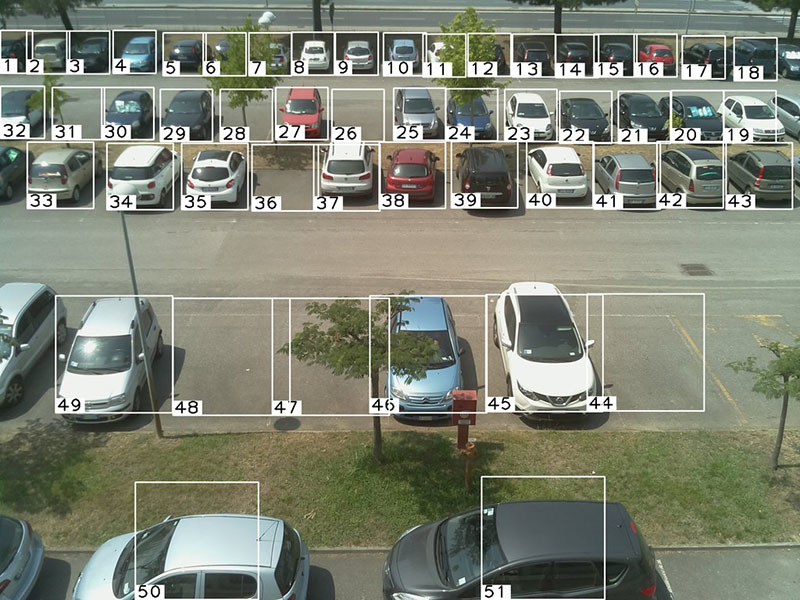}
    \includegraphics[width=50mm]{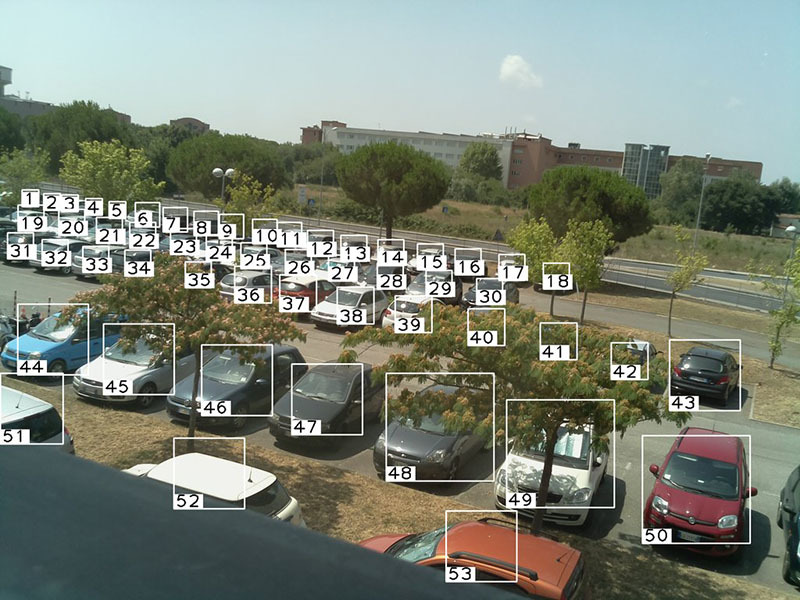}
    \caption{CNRPark A and B Dataset}
    \label{fig:5}
\end{figure}

This dataset \cite{amato2016car} consists of CNRPark A and CNRPark B and has 12,584 images (shown in Figure \ref{fig:5}). CNRPark A consists of 6,171 images taken using camera A, while CNRPark B contains 6,413 images captured using camera B. The dataset features image patches\footnote{cnrpark.it}  captured under varying lighting conditions, including those partially occluded by trees and shadowed by neighboring vehicles. This diverse composition evaluates the classifier's robustness under dynamic real-world conditions. The dataset evaluates a classifier's ability to adapt to varying camera perspectives. This is accomplished by training the model on images from one camera and evaluating its performance on images from a different camera. Details of the CNRPark dataset used in the experiments are shown in Table \ref{table:1}.
\begin{table}[h!!]
\caption{Details of CNRPark Patches dataset}
\begin{center}
\begin{tabular}{cccc}
\hline
Subsets & Bus\_patches & Free\_patches & Total\_patches \\
\hline
CNRPark A & 3622  & 2549 & 6171 \\
CNRPark B & 4781  & 1632 & 6413 \\
\hline
\end{tabular}
\label{table:1}
\end{center}
\end{table}

 \subsection{Custom Hazy Parking System dataset:}

 The dataset introduced by Satyanath et al. \cite{refpaper}, designed for classifying hazy images using a modified mAlexNet \cite{b10} model combined with AOD-Net \cite{AODNET}, consists of 752 original (unaugmented) image patches, including 502 hazy car patches and 250 haze-free patches. These patches are divided into distinct subsets, with 70\% allocated for training and 30\% reserved for testing. The test set maintains a balanced representation of occupied and unoccupied patches prior to augmentation.
To increase variability, image augmentation is performed using the Keras image preprocessing library\footnote{Chollet, François et al.,``Keras," https://keras.io, (2015).}. Post augmentation, the dataset grows to 5,010 labeled image patches, representing both occupied and unoccupied parking spaces under various hazy conditions, including pollution, snow, and with haze intensities ranging from mild to severe. Table \ref{table:2} provides detailed statistics of the dataset. 

\subsubsection{Improvisation in Custom Hazy Parking System Dataset}
\begin{figure}[h!!]
\begin{minipage}{0.19\textwidth}
\includegraphics[height=0.7in, width=0.95\textwidth]{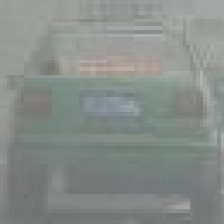}
\end{minipage}\hfill
\begin{minipage}{0.19\textwidth}
\includegraphics[height=0.7in, width=0.95\textwidth]{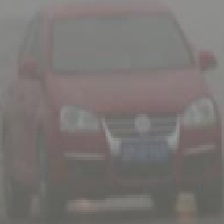}
\end{minipage}\hfill
\begin{minipage}{0.19\textwidth}
\includegraphics[height=0.7in, width=0.95\textwidth]{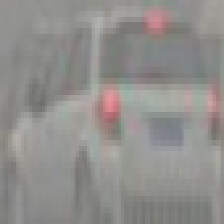}
\end{minipage}\hfill
\begin{minipage}{0.19\textwidth}
\includegraphics[height=0.7in, width=0.95\textwidth]{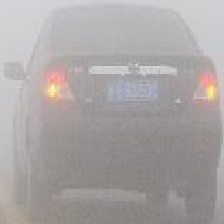}
\end{minipage}\hfill
\begin{minipage}{0.19\textwidth}
\includegraphics[height=0.7in, width=0.95\textwidth]{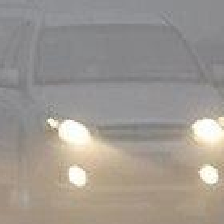}
\end{minipage}\hfill
\begin{minipage}{0.19\textwidth}
\includegraphics[height=0.7in, width=0.95\textwidth]{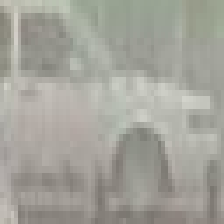}
\end{minipage}\hfill
\begin{minipage}{0.19\textwidth}
\includegraphics[height=0.7in, width=0.95\textwidth]{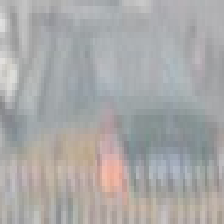}
\end{minipage}\hfill
\begin{minipage}{0.19\textwidth}
\includegraphics[height=0.7in, width=0.95\textwidth]{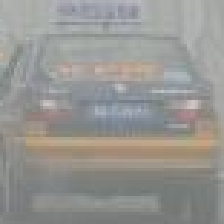}
\end{minipage}\hfill
\begin{minipage}{0.19\textwidth}
\includegraphics[height=0.7in, width=0.95\textwidth]{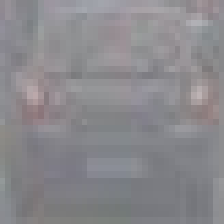}
\end{minipage}\hfill
\begin{minipage}{0.19\textwidth}
\includegraphics[height=0.7in, width=0.95\textwidth]{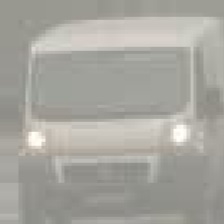}
\end{minipage}\hfill
\begin{minipage}{0.19\textwidth}
\includegraphics[height=0.7in, width=0.95\textwidth]{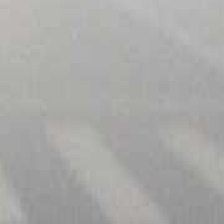}
\end{minipage}
\begin{minipage}{0.19\textwidth}
\includegraphics[height=0.7in, width=0.95\textwidth]{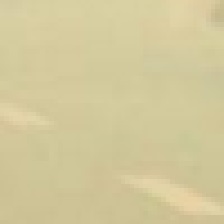}
\end{minipage}\hfill
\begin{minipage}{0.19\textwidth}
\includegraphics[height=0.7in, width=0.95\textwidth]{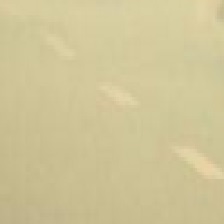}
\end{minipage}\hfill
\begin{minipage}{0.19\textwidth}
\includegraphics[height=0.7in, width=0.95\textwidth]{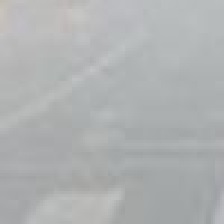}
\end{minipage}\hfill
\begin{minipage}{0.19\textwidth}
\includegraphics[height=0.7in, width=0.95\textwidth]{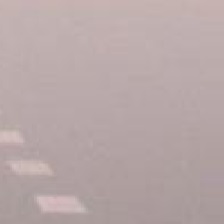}
\end{minipage}\hfill
\begin{minipage}{0.19\textwidth}
\includegraphics[height=0.7in, width=0.95\textwidth]{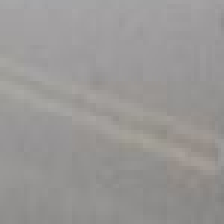}
\end{minipage}\hfill
\begin{minipage}{0.19\textwidth}
\includegraphics[height=0.7in, width=0.95\textwidth]{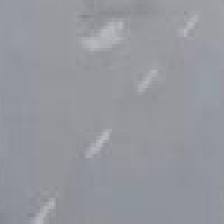}
\end{minipage}\hfill
\begin{minipage}{0.19\textwidth}
\includegraphics[height=0.7in, width=0.95\textwidth]{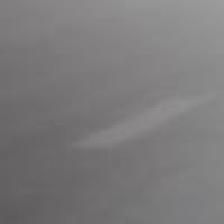}
\end{minipage}\hfill
\begin{minipage}{0.19\textwidth}
\includegraphics[height=0.7in, width=0.95\textwidth]{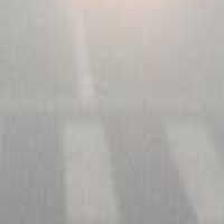}
\end{minipage}\hfill
\begin{minipage}{0.19\textwidth}
\includegraphics[height=0.7in, width=0.95\textwidth]{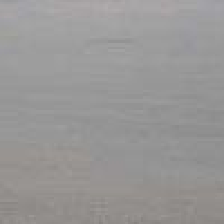}
\end{minipage}\hfill
\caption{Hazy Parking Dataset: Top two rows - busy parking space, Bottom two rows - free parking space}
\label{fig:parkingpatches}
\end{figure}

We observed that the hybrid model memorizes the patterns in augmentation when trained and tested on augmented data and performs extremely well. Hence, the dataset was altered such that the model was trained on a mix of unaugmented and augmented data and tested on an unseen unaugmented dataset of Hazy Images. This increases the robustness of the model and confidence in correct prediction with minimal learning and more generalization of patterns. Figure \ref{fig:parkingpatches} represents some busy and free parking space images of this custom dataset.

\begin{table}[H]
\caption{Sample images from Custom Hazy parking system dataset}
\begin{center}
\begin{tabular}{cccc}
\hline
Dataset & Free\_patches & Busy\_patches & Total\_patches \\
\hline
Hazy parking system & 2500 & 2510 & 5010 \\
Hazy parking system (Train) & 1500 & 1900 & 3400 \\
Hazy parking system (Test) & 1000 & 610 & 1610 \\ 
 \\
Unaugmented images & 250 & 502 & 752 \\ 
Unaugmented Train & 150 & 380 & 532 \\
Unaugmented Test & 100 & 122 & 222 \\

\hline
\end{tabular}
\label{table:2}
\end{center}
\end{table}

\section{Proposed system}

\subsection{Model Description}
We propose a hybrid model utilizing pre-trained feature extractors like ResNet-50, AlexNet, and GoogleNet and a Pin-GTSVM binary classifier. The proposed model is tested and compared  against a variety of models on different parking system datasets, highlighting its real-world applications and its inherent dehazing capabilities.

The Pin-GTSVM with RBF kernel (\texttt{pinGTSVM\_RBF}) is used to handle noisy and imbalanced data. This model incorporates the pinball loss function, which enhances its resilience to outliers and noise, and employs a RBF kernel to capture non-linear relationships in the data.
\newline

\subsection{Model Components}
\begin{itemize}
    \item \textbf{TestX}: A matrix where each row represents a test data point.
    \item \textbf{DataTrain}: A struct containing training data:
    \begin{itemize}
        \item \texttt{DataTrain.A}: Positive class training samples.
        \item \texttt{DataTrain.B}: Negative class training samples.
    \end{itemize}
    \item \textbf{FunPara}: A struct for model parameters:
    \begin{itemize}
        \item \texttt{FunPara.c1}: Regularization parameter for the first class.
        \item \texttt{FunPara.c2}: Regularization parameter for the second class.
        \item \texttt{FunPara.kerfPara}: Parameters for the kernel function.
        \item \texttt{FunPara.tau}: Parameter for the pinball loss function.
    \end{itemize}
\end{itemize}

Further, we have also compared the performance of the model using linear and  RBF Kernels.

\subsection{Optimization Problem}
The model solves two optimization problems to find two non-parallel hyperplanes:
\begin{itemize}
    \item For the positive class $A$:
    \[
    \min_{\mathbf{w}_1, b_1} \frac{1}{2} \|\mathbf{w}_1\|^2 + c_1 \sum_{i=1}^{m_2} L_{\tau}(\mathbf{w}_1^T \phi(\mathbf{x}_i^B) + b_1)
    \]
    \item For the negative class $B$:
    \[
    \min_{\mathbf{w}_2, b_2} \frac{1}{2} \|\mathbf{w}_2\|^2 + c_2 \sum_{i=1}^{m_1} L_{\tau}(\mathbf{w}_2^T \phi(\mathbf{x}_i^A) + b_2)
    \]
\end{itemize}
Here, $\phi(\cdot)$ represents the feature mapping induced by the RBF kernel, and $c_1, c_2$ are regularization parameters.

\subsection{Model Outputs}
Following are the descriptions of the variables used in the algorithm for recording performance:
\begin{itemize}
    \item \textbf{Predict\_Y}: The predicted labels for the test data.
    \item \textbf{A, B}: Training data for positive and negative classes.
    \item \textbf{w1, b1}: Weight vector and bias term for the positive class hyperplane.
    \item \textbf{w2, b2}: Weight vector and bias term for the negative class hyperplane.
    \item \textbf{acc}: The model's accuracy on the test data.
    \item \textbf{err}: The model's error rate on the test data.
    \item \textbf{time1}: The computational time is taken to train the model and make predictions.
\end{itemize}

\subsection{Implementation Details:}

The implementation of \texttt{pinGTSVM\_RBF} involves:

\begin{itemize}
    \item Kernel Matrix Computation: Both Linear and Gaussian kernels for training and testing datasets.
    \item Optimization via Quadratic Programming: Solving the quadratic programming problems to obtain the weight vectors ($w1, w2$) and bias terms ($b1, b2$) for the twin hyperplanes. We use the \texttt{quadprog} in \texttt{MATLAB} to solve these QPPs
\end{itemize}

The model thus effectively integrates the pinball loss function with a robust twin SVM framework, providing enhanced noise and outlier resistance. 

\subsection{Algorithm Analysis}

\begin{figure}[h!!]
    \centering
    \captionsetup{width=\textwidth}
    \includegraphics[height=6in, width=0.80\textwidth]{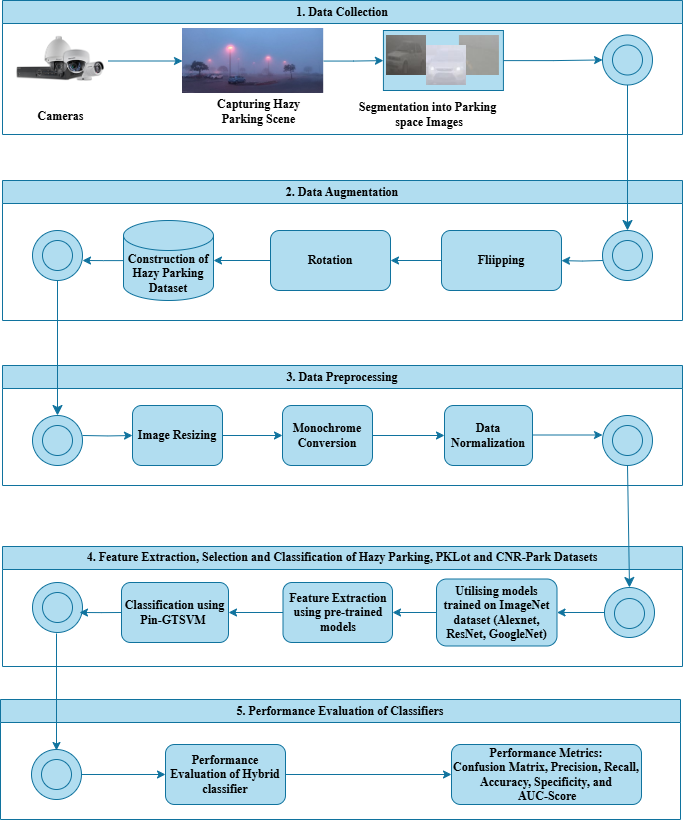}
    \caption{Working diagram of the proposed model for Parking Space Classification under hazy conditions}
    \label{fig:7}
\end{figure}

A simple working diagram of the proposed model is shown in Figure \ref{fig:7}  to classify the parking slot as a busy or free.

\subsection{Feature Extraction}
In this study, we explored multiple pre-trained models of CNN architectures to extract the features for classifying parking space occupancy. Alongside ResNet-50, we utilized AlexNet and GoogLeNet. Each model has distinct characteristics that contribute to their ability to capture useful features from images. We selected specific layers from each model for feature extraction, such as for Res-Net-50, the fc1000 layer was used, AlexNet utilized the fc7 layer, and GoogLeNet used its loss3-classifier layer.

We defined a generalized function \texttt{extract features} that can handle these pre-trained CNN models by taking an additional parameter for the network type. To extract features, we employed a pre-trained ResNet-50 network, focusing on the fc1000 layer. Images were processed in batches to optimize memory usage, resizing each to 224×224 pixels to match the network input requirements. GPU acceleration was utilized where available to expedite computation. The activations function extracted 1000-dimensional feature vectors for each image, which were aggregated into a feature matrix. This matrix served as input for subsequent classification tasks, ensuring efficient and scalable feature extraction while maintaining compatibility with large datasets.

\subsection{Classification}
This section elucidates the methodology employed for feature extraction and subsequent classification using Pin-GTSVM with both RBF and linear kernels. The classification results obtained demonstrate its efficacy in accurately categorizing parking lot images into busy and free classes. The optimized Pin-GTSVM model achieves high accuracy, showcasing the effectiveness of combining deep learning-based feature extraction with advanced classification techniques. Additionally, the comparison between RBF and linear kernels highlights the importance of selecting appropriate kernel functions based on the dataset's characteristics.

\subsection{Parameter Selection and Tuning}
As noted in \cite{pingtsvm}, the performance of various algorithms is significantly influenced by the choice of parameters. In the experimental work, the optimal parameter values were determined using the grid search \footnote{https://scikit-learn.org/1.5/modules/grid\_search.html} method. All algorithms require the selection of three key parameters: the penalty parameters c1, c2, and Gaussian kernel parameter \textmu. Three values of $\tau$1, $\tau$2 are selected as 0.5, 0.8, and 1. For the penalty parameters c1 and c2, the optimal values are selected from the set \(\{2^{-5}, 2^{-4}, \ldots, 2^4, 2^5\}\). Additionally, the parameter \(\mu\) was chosen from the set \(\{2^{-10}, 2^{-9}, \ldots, 2^9, 2^{10}\}\). This range of values allows for a comprehensive search of parameter space to optimize the performance of the proposed model.
\newline
 
 For reducing the computation cost of parameter selection, we set $\tau$1 = $\tau$2. The optimal set of parameters is identified and presented, along with the highest accuracy achieved for those parameters. A range of accuracies is observed across various parameter combinations, reflecting the variation in results and the shifting of hyperplanes during classification. All the models were trained on MATLAB 2023b in a Windows 11 OS with 32 GB RAM running on Intel(R) Core(TM) i7-9750H CPU @ 2.60GHz in CPU Mode.

\section{Result Analysis}
We used different pre-trained models on the ImageNet dataset for feature extraction and compared the changes in accuracy. A second type of comparison was done on the basis of different kernels used and at different quantile levels, i.e., in the lower quartile, inter-quartile range, and the upper quartile.

The earlier State-of-The-Art AOD-NET joint optimized model \ref{table:6} trained on hazy images of the OTS dataset achieved a maximum accuracy of 88.39\% on the "Hazy Parking System" dataset. The proposed model clearly performs better with a maximum accuracy of 98.20\%. The details of the modified Alexnet model can be found in Table \ref{table:3}

\begin{table}[H]
\caption{Modified mAlexnet architecture}
\begin{tabular}{|C{3cm}|C{3cm}|C{3cm}|C{2.625cm}|C{2.625cm}|}
\hline
Conv1 & Conv2 & Conv3 & Fc4 & Fc5 \\
\hline
IS: 224x224x3 & IS: 16x27x27 & IS: 20x11x11 & IS: 480 & IS : 48 \\ 
16x11x11+4 & 20x5x5+1 & 30x3x3+1 & - & - \\
pool 3x3+2 & pool 3x3+2 & pool 3x3+2 & 48 & 2 \\
Batchnorm & Batchnorm & - & - & - \\
ReLU & ReLU & ReLU & ReLU & Softmax \\
\hline
\end{tabular}
\label{table:3}
\end{table}

\begin{table}[H]
\caption{ Effects of different pre-trained Feature Extractor Models on Accuracy }
\begin{center}
\begin{tabular}{ccccc}
\hline
Datasets & RESNET-50 & AlexNet & GoogleNet & modified mAlexnet \cr
\hline
Hazy parking system & \textbf{98.20} & 97.75 & 96.83 & 74.41 \\ \\
CNRPark A (Train) and B (Test) & \textbf{89.51} & 87.98 & 87.93 & 87.56 \\ \\
CNRPark B (Train) and A (Test) & \textbf{98.70} & 98.35 & 97.36 & 90.78 \\ \\
\hline
\end{tabular}
\label{table:4}
\end{center}
\end{table}

 Table \ref{table:4} shows the robustness of the proposed model in every scenario with or without Atmospheric Noise, as we have very low standard deviations from the best mean accuracies obtained in cross-validation.

\begin{table}[H]
\caption{ 5-fold Cross-Validation Mean Accuracies and Standard Deviations }
\begin{center}
\begin{tabular}{cccc}
\hline
Datasets & RESNET-50 & AlexNet & GoogleNet  \cr
\hline
Hazy parking system & \textbf{99.85(-0.28)} & 99.77(-0.11) & 99.75(-0.09)  \\ \\
CNRPark A (Train) and B (Test) & 99.53(-0.11) & \textbf{99.98(-0.04)} & 99.76(-0.09)  \\ \\
CNRPark B (Train) and A (Test) & 96.73(-0.41) & \textbf{98.75(-0.38)} & 97.8(-0.37)  \\ \\
\hline 
\end{tabular}
\label{table:5}
\end{center}
\end{table}

\textit{Note}: To verify the claim stated above, the average cross-validation accuracy (99.98\%) for AlexNet on CNRPark A(Train) and B(Test) shown in Table \ref{table:5}, and low standard deviation of (0.04\%) indicate that the proposed model is robust and not overfitting excessively to the training data. We evaluate and compare the performance of the models presented in \cite{refpaper}, \cite{b10}, and \cite{b11} on the Hazy Parking System Dataset. Model architectures of  1, 2, 3, and modified mAlexnet are taken from the paper \cite{refpaper}

The joint optimization significantly improves the performance on hazy images in most cases but consistently reduces accuracy under non-hazy conditions. Nevertheless, as demonstrated in Table \ref{table:6}, the proposed hybrid model outperforms all other models by a significant margin.

\begin{table}[h!!]
\caption{Accuracy comparison of models on the hazy parking system test dataset}
\begin{tabular}{C{10cm}C{5cm}}
\hline
Architectures & Accuracy on Hazy Parking System Test   \\
\hline
Model 1 \cite{refpaper} & 86.71\% \\
Model 1 after joint optimization & 88.39\%  \\ \\
Model 2 \cite{refpaper}& 86.40\% \\
Model 2 after joint optimization & 86.89\% \\ \\ 
Model 3 ($\lambda$=0.1916) \cite{refpaper} & 86.27\% \\
Model 3 ($\lambda$=0.1916) after joint optimization & 88.07\%  \\ \\
Model 3 ($\lambda$=0.4290) & 86.89\% \\ 
Model 3 ($\lambda$=0.4290) after joint optimization & 88.63\%  \\ \\
Model 3 ($\lambda$=0.6554) & 86.96\% \\
Model 3  ($\lambda$=0.6554) after joint optimization & 86.27\%  \\ \\
Model 3 ($\lambda$=0.8984) & 84.78\% \\
Model 3 ($\lambda$=0.8984) after joint optimization & 86.71\%  \\ \\
Modified mAlexnet \cite{refpaper} & 74.41\% \\ \\
mAlexnet\cite{b10} & 76.27\% \\ \\
Alexnet\cite{b11} & 80.37\% \\ \\
Hybrid ResNet50 - PinGTSVM & \textbf{98.20}\% \\ \\
Hybrid GoogleNet - PinGTSVM & 97.75\% \\ \\
Hybrid AlexNet - PinGTSVM & 96.83\% \\ \\
\hline
\end{tabular}
\label{table:6}
\end{table}

\subsection{Significance of the Parameter \(\tau\) in the proposed Model}

The parameter \(\tau\) plays a crucial role in the performance and behavior of the proposed model. Depending on the specific context and algorithm used, \(\tau\) has various implications. In this study, we systematically explored the effects of different \(\tau\) values to identify its optimal setting and understand its influence on model performance. In Table \ref{table:7}, we outline the significance of \(\tau\) the proposed model and its empirical impact based on the findings.\newline

\begin{table}[h!!]
\caption{Effects of different values of $\tau$ (Asymmetry) on Best Accuracies of Classification Models}
\begin{center}
\label{table:7}
\resizebox{\textwidth}{!}{
\begin{tabular}{|l|l|l|l|l|l|l|l|l|l|l|l|l|}
\hline
\multirow{3}{*}{Dataset/$\tau$} &
\multicolumn{3}{c|}{RESNET-50(\%)} &
\multicolumn{3}{c|}{AlexNet(\%)} &
\multicolumn{3}{c|}{GoogleNet(\%)} \\
\cline{2-10}
& 0.5 & 0.8 & 1 & 0.5 & 0.8 & 1 & 0.5 & 0.8 & 1 \\
\hline
\hline
Hazy parking system & \textbf{98.20} & \textbf{98.20} & 95.95 & \textbf{97.75} & 97.28 & 95.50 & 95.95 & \textbf{96.83} & 95.50 \\
\hline
CNRPark A (Train) and B (Test) & 89.44 & \textbf{89.51} & 87.23 & \textbf{87.98} & 87.90 & 87.23 & 85.79 & \textbf{87.93} & 82.00 \\
\hline
CNRPark B (Train) and A (Test) & \textbf{98.70} & 98.35 & 97.18 & 97.47 & \textbf{98.35} & 83.73 & 97.02 & \textbf{97.36} & 87.41 \\
\hline
\end{tabular}
}
\end{center}

\end{table}

As stated earlier, in the proposed model, \(\tau\) acts as a regularization parameter that balances the trade-off between the fitting of the training data closely and maintaining model simplicity to prevent overfitting. The following observations were made:
\begin{itemize}
    \item \textbf{Low \(\tau\)}: When \(\tau\) is set to a low value, the model tends to fit the training data more closely, capturing even minor fluctuations. This can lead to overfitting, where the model performs well on training data but poorly on unseen testing data. Hence, lower values of \(\tau)\), i.e., less than 0.5, are neglected.
    \item \textbf{High \(\tau\)}: Increasing \(\tau\) simplifies the model by penalizing complexity. This helps prevent overfitting and improves the model’s generalization to new data. However, setting \(\tau\) too high can result in underfitting, where the model is too simple to capture the underlying patterns in the data.
\end{itemize}

\subsubsection{Sensitivity Analysis}

We conducted a sensitivity analysis to explore the impact of varying \(\tau\)   on model performance as shown in Table \ref{table:7}.
\begin{itemize}
    \item \textbf{Performance Consistency}: The model showed robust performance for \(\tau\) values in the vicinity of 0.5, with slight variations in accuracy. This indicates that while \(\tau = 0.5\) is optimal, the model is not overly sensitive to minor changes in \(\tau\) as observed at a value equal to 0.8.
    \item \textbf{Performance Peaks and Drops}: Significant deviations from the optimal \(\tau\) value (both lower and higher) resulted in noticeable drops in accuracy, underscoring the importance of proper tuning.
\end{itemize}

\subsubsection{Practical Implications}

The choice of \(\tau\) has practical implications for model deployment:
\begin{itemize}
    \item \textbf{Model Complexity}: Proper tuning of \(\tau\) ensures an optimal balance between model complexity and generalization. This is crucial for real-world applications where overfitting can lead to poor performance on new data.
    \item \textbf{Consistency and Reliability}: A well-tuned \(\tau\) contributes to consistent and reliable model performance, which is essential for maintaining trust in predictive analytics.
\end{itemize}

\subsection{Comparison based on Linear and Gaussian Kernels}

\begin{table}[H]
\caption{ Best Accuracy using Linear Kernel and Gaussian Kernel (Non-Linear) }
\begin{center}
\begin{tabular}{ccc}
\hline
Datasets & Linear & Gaussian \\
\hline
Hazy parking system & 64.41 & \textbf{98.20} \\ \\
CNRPark A (Train) and B (Test) & 46.98 & \textbf{89.51} \\ \\
CNRPark B (Train) and A (Test) & 50.30 & \textbf{98.70}  \\ \\
\hline
\end{tabular}
\label{table:8}
\end{center}
\end{table}

Linear kernels are computationally efficient and suitable for high-dimensional data where linear separability is achievable. In contrast, Gaussian kernels (Non-Linear Kernels) can handle more complex patterns and interactions within the data, making them ideal for datasets with non-linear relationships. The choice between linear and Gaussian kernels depends on the specific nature of the data and the problem shown in Table \ref{table:8}.

These kernel functions enable SVMs and other algorithms to transform data into higher-dimensional spaces, facilitating the separation of classes and regression tasks in both linearly and non-linearly separable cases. We used Non-Linear Kernel because the data obtained from the Hazy parking system dataset is very complex in nature (i.e., not easily separable) and, hence, difficult to classify with the help of a Linear Kernel. This inference is observed from the Table \ref{table:8}.

\subsection{Comparison based on Confusion Matrix}
A confusion Matrix is a fundamental tool used to evaluate Machine Learning models. It summarizes a classification model's performance by presenting the counts of True Positives (TP), False Positives (FP), True Negatives (TN), and False Negatives(FN), comparing the model's predictions to the actual outcomes in the test dataset.

\begin{figure}[ht]
    \centering
    \begin{minipage}{0.3\linewidth}
        \centering
        \includegraphics[width=\linewidth]{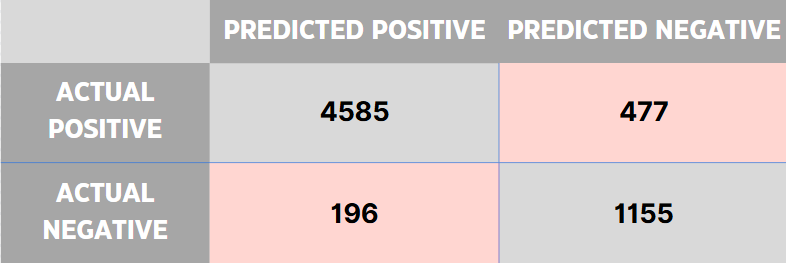}
        \subcaption{CNR Park A}
    \end{minipage}
    \hfill
    \begin{minipage}{0.3\linewidth}
        \centering
        \includegraphics[width=\linewidth]{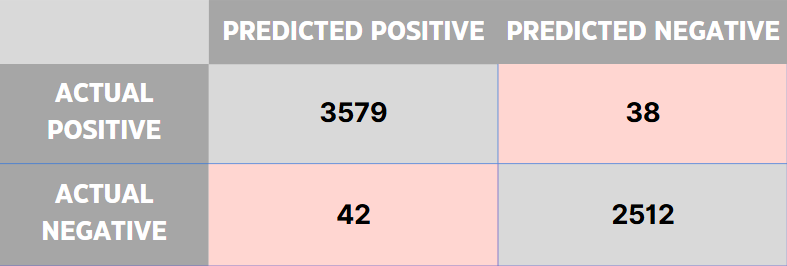}
        \subcaption{CNR Park B}
    \end{minipage}
    \hfill
    \begin{minipage}{0.3\linewidth}
        \centering
        \includegraphics[width=\linewidth]{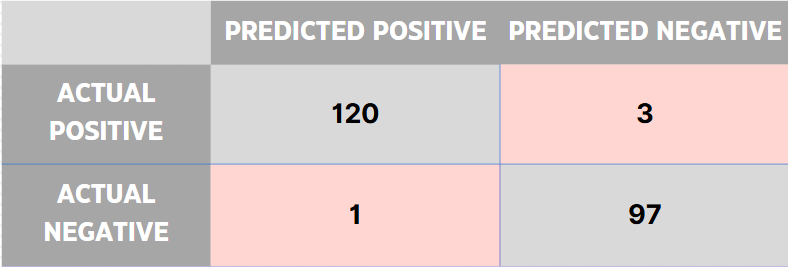}
        \subcaption{Hazy Parking}
    \end{minipage}
    \caption{ResNet-50}
    \label{fig:8}
\end{figure}

\begin{figure}[ht]
    \centering
    \begin{minipage}{0.3\linewidth}
        \centering
        \includegraphics[width=\linewidth]{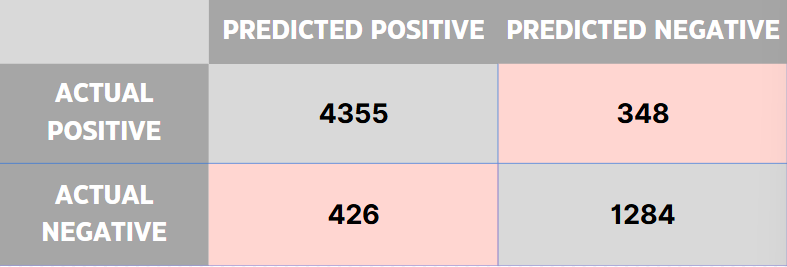}
        \subcaption{CNR Park A}
    \end{minipage}
    \hfill
    \begin{minipage}{0.3\linewidth}
        \centering
        \includegraphics[width=\linewidth]{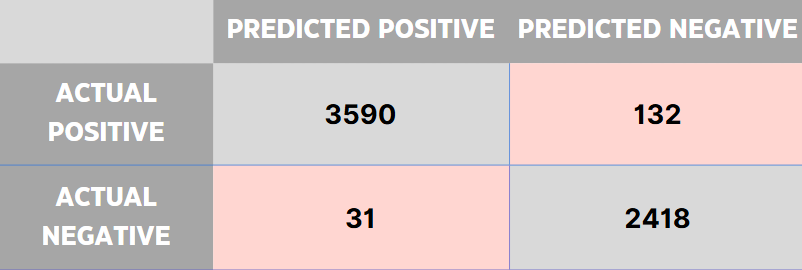}
        \subcaption{CNR Park B}
    \end{minipage}
    \hfill
    \begin{minipage}{0.3\linewidth}
        \centering
        \includegraphics[width=\linewidth]{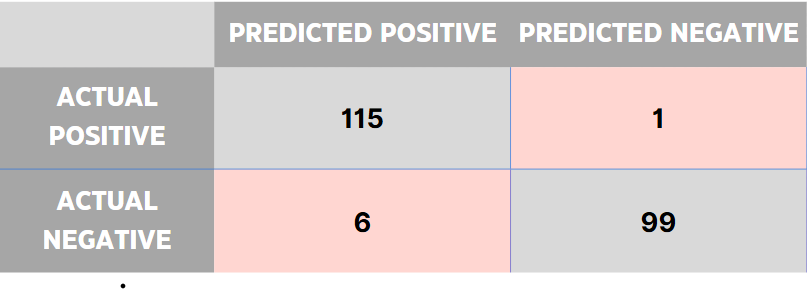}
        \subcaption{Hazy Parking}
    \end{minipage}
    \caption{GoogleNet}
    \label{fig:9}
\end{figure}

\begin{figure}[ht]
    \centering
    \begin{minipage}{0.3\linewidth}
        \centering
        \includegraphics[width=\linewidth]{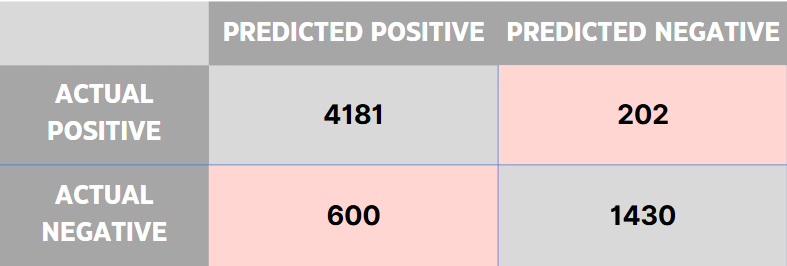}
        \subcaption{CNR Park A}
    \end{minipage}
    \hfill
    \begin{minipage}{0.3\linewidth}
        \centering
        \includegraphics[width=\linewidth]{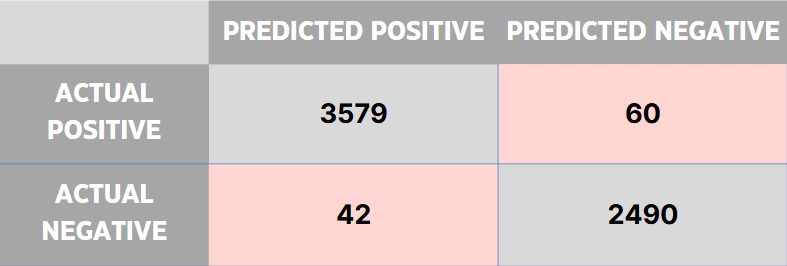}
        \subcaption{CNR Park B}
    \end{minipage}
    \hfill
    \begin{minipage}{0.3\linewidth}
        \centering
        \includegraphics[width=\linewidth]{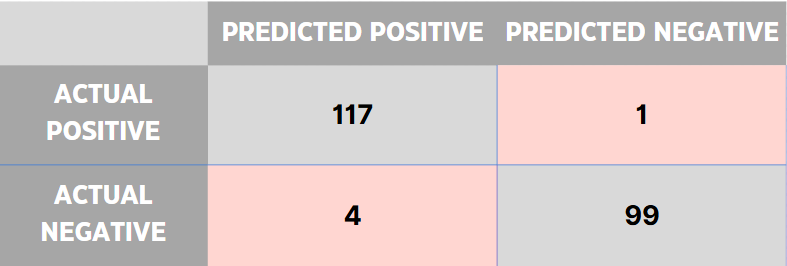}
        \subcaption{Hazy Parking}
    \end{minipage}
    \caption{AlexNet}
    \label{fig:10}
\end{figure}

Figures \ref{fig:8}, \ref{fig:9}, and \ref{fig:10} represent the confusion matrices of ResNet-50, GoogleNet, and AlexNet trained on CNR Park A, CNR Park B, and Hazy Parking datasets, respectively. 

We extract the counts of TP, FP, TN, and FN from all the confusion matrices and calculate the following performance metrics. A brief about the metrics utilized to benchmark results obtained is stated below:\newline

Precision is used to evaluate the accuracy of positive predictions made by a model.
\[
\text{Precision} = \frac{\text{TP}}{\text{TP} + \text{FP}}
\]

Recall measures the ability of a classifier to identify positive instances correctly.
\[
\text{Recall} = \frac{\text{TP}}{\text{TP} + \text{FN}}
\]

The F1 score is the harmonic mean of precision and recall.

Specificity measures the proportion of actual negative cases correctly identified.
\[
\text{Specificity} = \frac{\text{TN}}{\text{TN} + \text{FP}}
\]

\begin{table}[H]
\caption{ Precision Values Comparison of Different Architectures: }
\begin{center}
\begin{tabular}{cccc}
\hline
Datasets & RESNET-50 & AlexNet & GoogleNet  \cr
\hline
Hazy parking system & 0.98 & \textbf{0.99} & \textbf{0.99}  \\ \\
CNRPark A (Train) and B (Test) & 0.91 & \textbf{0.95} & 0.93  \\ \\
CNRPark B (Train) and A (Test) & \textbf{0.99} & 0.98 & 0.96  \\ \\
\hline
\end{tabular}
\label{table:9}
\end{center}
\end{table}

As seen in the Table \ref{table:9}, ResNet-50, AlexNet, and GoogleNet, all demonstrate state-of-the-art performance in dealing with hazy images when coupled with a Pin-GTSVM Classifier, keeping in mind their architectural differences and handling of fine-grained features affected by haze.

\begin{table}[H]
\caption{ Recall Values Comparison of Different Architectures: }
\begin{center}
\begin{tabular}{cccc}
\hline
Datasets & RESNET-50 & AlexNet & GoogleNet  \cr
\hline
Hazy parking system & \textbf{0.98} & 0.97 & 0.95  \\ \\
CNRPark A (Train) and B (Test) & \textbf{0.96} & 0.87 & 0.91  \\ \\
CNRPark B (Train) and A (Test) & \textbf{0.99} & \textbf{0.99} & \textbf{0.99}  \\ 
\hline
\end{tabular}
\label{table:10}
\end{center}
\end{table}

From the results shown in Table \ref{table:10}, RESNET-50 consistently achieves high recall across different datasets compared to AlexNet and GoogleNet. Particularly, in the ``Hazy parking system" and ``CNRPark A (Train) and B (Test)" datasets, RESNET-50 demonstrates superior performance with recall values of 0.98 and 0.96, respectively, highlighting its effectiveness in these contexts. This also suggests that the ResNet-50 and Pin-GTSVM hybrid models can be superior when high recall rates are crucial.

\begin{table}[H]
\caption{ F-1 Scores Comparison of Different Architectures: }
\begin{center}
\begin{tabular}{cccc}
\hline
Datasets & RESNET-50 & AlexNet & GoogleNet  \cr
\hline
Hazy parking system & \textbf{0.98} & \textbf{0.98} & 0.97  \\ \\
CNRPark A (Train) and B (Test) & \textbf{0.93} & 0.91 & 0.92  \\ \\
CNRPark B (Train) and A (Test) & \textbf{0.99} & \textbf{0.99} & 0.98  \\ \\
\hline
\end{tabular}
\label{table:11}
\end{center}
\end{table}

From the results shown in Table \ref{table:11}, RESNET-50 once again capitalizes and consistently achieves competitive F-1 scores across different datasets compared to AlexNet and GoogleNet. Particularly noteworthy is its performance in the ``Hazy parking system" and ``CNRPark B (Train) and A (Test)" datasets, where it achieves F-1 scores of 0.98 and 0.99, respectively, indicating robust performance in varied conditions.

\begin{table}[H]
\caption{ Specificity Comparison of Different Architectures: }
\begin{center}
\begin{tabular}{cccc}
\hline
Datasets & RESNET-50 & AlexNet & GoogleNet  \cr
\hline
Hazy parking system & 0.97 & \textbf{0.99} & \textbf{0.99}  \\ \\
CNRPark A (Train) and B (Test) & 0.71 & \textbf{0.88} & 0.79  \\ \\
CNRPark B (Train) and A (Test) & \textbf{0.99} & 0.98 & 0.95  \\ \\
\hline
\end{tabular}
\label{table:12}

\end{center}
\end{table}

AlexNet and GoogleNet consistently achieve higher specificity values than RESNET-50 across different datasets shown in Table \ref{table:12}. Notably, AlexNet shows superior specificity in the ``Hazy parking system" and ``CNRPark A (Train) and B (Test)" datasets. At the same time, RESNET-50 performs better in specific scenarios such as ``CNRPark B (Train) and A (Test)," where it achieves a specificity of 0.99. CNRPark B might have more complex or distinctive features compared to A, favoring ResNet-50's deeper architecture for capturing fine-grained details. AlexNet's slightly lower specificity indicates its limited capacity to handle the increased complexity of features from B to A.

\newpage
\subsection{False Positive Predictions}
This section discusses the number of false positive images predicted by ResNet, GoogleNet, and AlexNet models.
\begin{figure}[ht]
    \centering
    \begin{minipage}{0.25\linewidth}
        \centering
        \includegraphics[width=\linewidth]{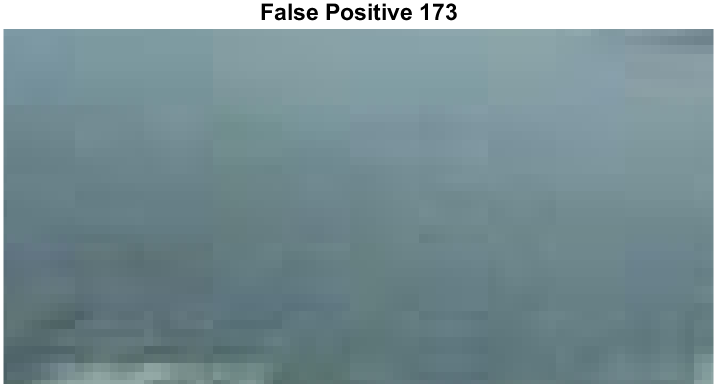}
    \end{minipage}
    \hfill
    \begin{minipage}{0.27\linewidth}
        \centering
        \includegraphics[width=\linewidth]{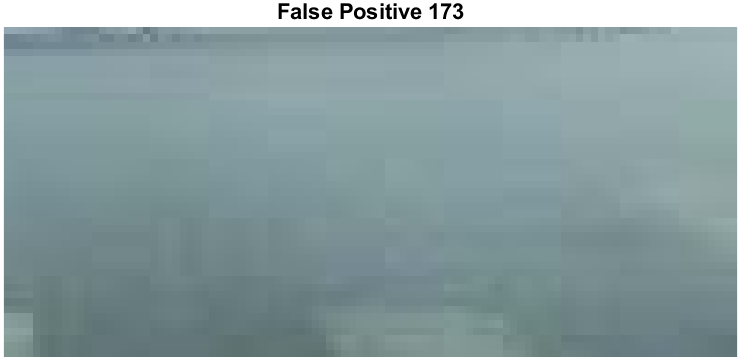}
    \end{minipage}
    \hfill
    \begin{minipage}{0.16\linewidth}
        \centering
        \includegraphics[width=\linewidth]{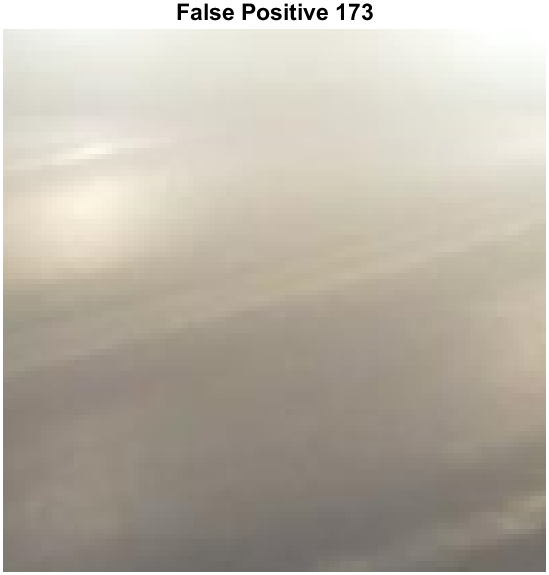}
    \end{minipage}
    \label{fig:12}
\end{figure}

The ResNet model predicted the above three images as false Positives.

\begin{figure}[H]
    \centering
    \includegraphics[width=50mm]{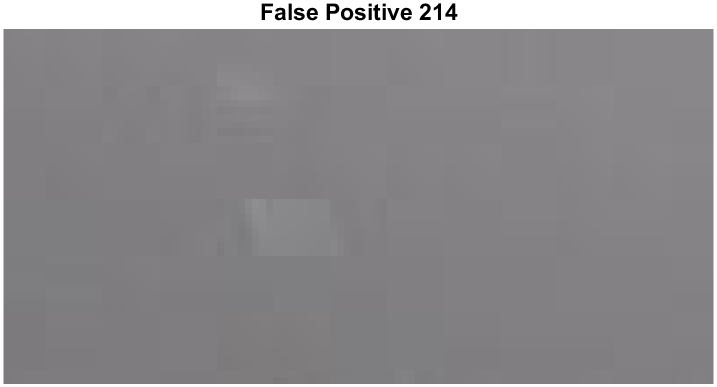}
    \label{fig:13}
\end{figure}

The GoogleNet Model predicted only one image as a false positive.

\begin{figure}[ht]
    \centering
    \begin{minipage}{0.24\linewidth}
        \centering
        \includegraphics[width=\linewidth]{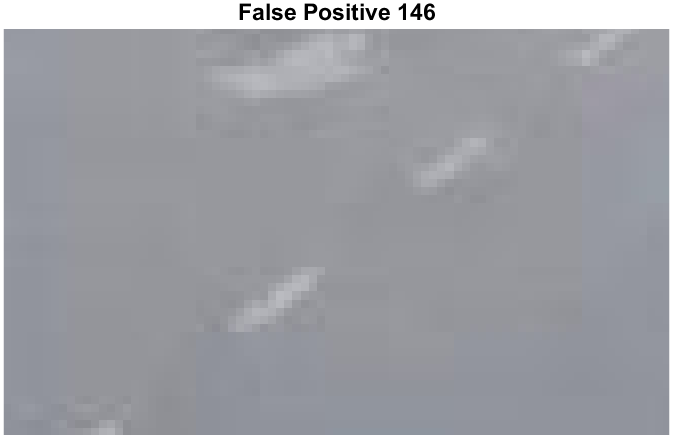}
    \end{minipage}
    \hfill
    \begin{minipage}{0.27\linewidth}
        \centering
        \includegraphics[width=\linewidth]{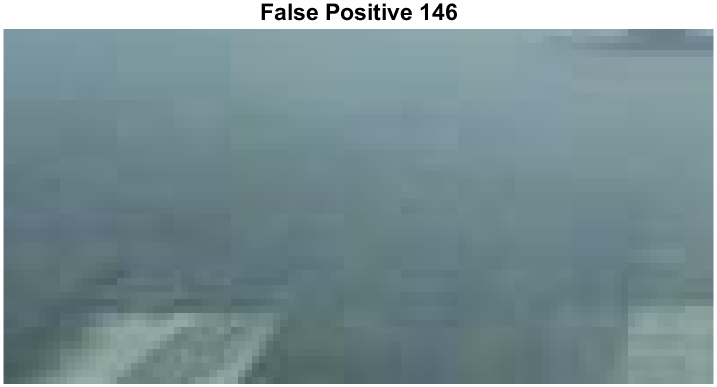}
    \end{minipage}
    \hfill
    \begin{minipage}{0.18\linewidth}
        \centering
        \includegraphics[width=\linewidth]{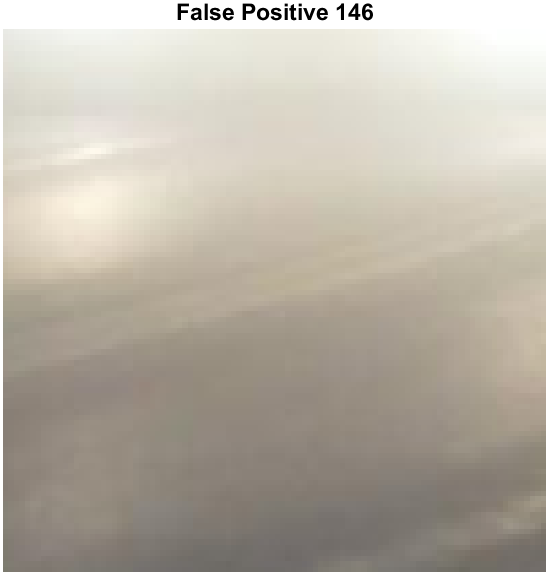}
    \end{minipage}
    \label{fig:14}
\end{figure}

The AlexNet model predicted three false positive images.

\subsection{Runtime Analysis}

\begin{table}[H]
\caption{ Pin-GTSVM Classification Runtime(in seconds) Comparison of Different Architectures }
\begin{center}
\begin{tabular}{cccc}
\hline
Datasets & RESNET-50 & AlexNet & GoogleNet  \cr
\hline
Hazy parking system & 43.52 & 47.88 & \textbf{38.65}  \\ \\
CNRPark A (Train) and B (Test) & \textbf{97.12} & 130.96 & 106.81 \\ \\
CNRPark B (Train) and A (Test) & 117.72 & 179.45 & \textbf{111.98}  \\ \\
\hline
\end{tabular}
\label{table:13}
\end{center}
\end{table}

GoogleNet demonstrates the shortest classification runtime for test images across all datasets shown in Table \ref{table:13}, indicating its efficiency in processing tasks with Pin-GTSVM. GoogLeNet, although deeper than AlexNet, achieves its depth with fewer parameters due to its modular design, and RESNET-50 involves a much greater number of layers, which can increase inference time compared to the relatively shallower GoogLeNet. However, it shows competitive performance in certain scenarios, such as ``CNRPark A (Train) and B (Test)" with a runtime of 97.12 seconds.

\subsection{Comparison of architectures on PKLot Dataset}
This dataset classifies each segmented parking space image as `occupied’ or `empty.’ These cropped images are organized into three directories: UFP04, UFP05, and PUCPR. The first two directories include images taken from different angles, specifically from the fourth and fifth floors of the UFPR building. In contrast, the third directory contains images captured from the tenth floor of the administration building at PUCPR. This arrangement ensures a diverse representation of the parking area, incorporating views from multiple elevations and positions.\\

We have trained and tested the proposed model (ResNet-50 and Pin-GTSVM) on the PKLot dataset and recorded the results shown in Table \ref{table:14}.
\begin{table}[H]

\footnotesize
\caption{Results for PKLot dataset}
\begin{center}

\begin{tabular}{cccccc}
\hline
Datasets & Precision & Recall & F1-Score & Specificity & Runtime (in sec) \cr
\hline
PUCPR(Train) and UFPR04(Test) & 0.9055 & 0.996 & 0.9486 & 0.896 & 50.7005  \\ \\
PUCPR(Train) and UFPR05(Test) & 0.9521 & 0.994 & 0.9726 & 0.95 & 51.0132  \\ \\
UFPR05(Train) and UFPR04(Test) & 0.9921 & 0.998 & 0.995 & 0.992 & 59.4975  \\ \\
UFPR04(Train) and UFPR05(Test) & 0.9802 & 0.988 & 0.9841 & 0.98 & 59.8688  \\ \\
UFPR04(Train) and PUCPR(Test) & 0.998 & 0.996 & 0.997 & 0.998 & 44.8916  \\ \\
UFPR05(Train) and PUCPR(Test) & 0.9784 & 0.994 & 0.9861 & 0.978 & 49.747 \\ \\
\hline
\end{tabular}
\label{table:14}
\end{center}
\end{table}

Table \ref{table:15} demonstrates the accuracy comparison of model against variants of AlexNet based models across the PKLot datasets. As can be clearly seen in Table \ref{table:15}, the proposed model performs better than the existing State-of-The-Art models.

\begin{table}[H]
\caption{Accuracy comparison of different AlexNet-based models with proposed model}
\centering
\begin{tabular}{llccc}
\hline
\textbf{Model} & \textbf{Training} & \multicolumn{3}{c}{\textbf{Testing Accuracy (\%)}} \\
\cline{3-5}
& & UFPR04 & UFPR05 & PUCPR \\
\hline
\multirow{3}{*}{AlexNet-SVM} & UFPR04 & 97.77 & 95.66 & 96.88 \\
& UFPR05 & 96.34 & 97.32 & 95.09 \\
& PUCPR & 98.21 & 96.45 & 99.17 \\
\hline
\multirow{3}{*}{AlexNet-LR} & UFPR04 & 96.36 & 97.33 & 95.06 \\
& UFPR05 & 97.71 & 95.62 & 96.82 \\
& PUCPR & 98.17 & 96.42 & 99.14 \\
\hline
\multirow{3}{*}{AlexNet-RF} & UFPR04 & \textbf{98.92} & 95.62 & 98.39 \\
& UFPR05 & 97.77 & 98.45 & 96.99 \\
& PUCPR & \textbf{99.03} & 97.87 & 99.58 \\
\hline
\multirow{3}{*}{AlexNet-GNB} & UFPR04 & 96.62 & 93.86 & 95.69 \\
& UFPR05 & 95.14 & 96.44 & 93.66 \\
& PUCPR & 96.37 & 94.83 & 97.76 \\
\hline
\multirow{3}{*}{AlexNet-KNN} & UFPR04 & 98.68 & 96.80 & 97.68 \\
& UFPR05 & 96.78 & 97.83 & 95.79 \\
& PUCPR & 98.70 & 94.83 & 99.38 \\
\hline
\multirow{3}{*}{AlexNet-XGBoost} & UFPR04 & 98.68 & 96.80 & 97.68 \\
& UFPR05 & 96.78 & 97.83 & 95.79 \\
& PUCPR & 98.70 & 94.83 & 99.38 \\
\hline
\multirow{3}{*}{AlexNet-MLELM} & UFPR04 & 96.76 & 93.58 & 95.40 \\
& UFPR05 & 95.70 & 96.69 & 93.50 \\
& PUCPR & 98.92 & \textbf{98.25} & \textbf{99.85} \\
\hline
\multirow{3}{*}{ResNet-PinGTSVM} & UFPR04 & - & \textbf{98.40} & \textbf{99.70} \\
 & UFPR05 & \textbf{99.50} & - & \textbf{98.60} \\
 & PUCPR & 94.60 & 97.20 & - \\
\hline
\end{tabular}
\label{table:15}
\end{table}

Table \ref{table:16} draws a comparison with other widely used architectures such as VGG19, the base model(ResNet50), CarNet etc. and some newer State-of-The-Art architectures such as APSD-OC.
\begin{table}[H]
\caption{Comparison of different pre-trained models' accuracy across datasets}
\footnotesize
\centering
\begin{tabular}{llllccc}
\hline
\textbf{Author(s)} & \textbf{Model} & \textbf{Training} & \multicolumn{3}{c}{\textbf{Accuracy (\%)}} \\
\cline{4-6}
& & & UFPR04 & UFPR05 & PUCPR \\
\hline
\multirow{3}{*}{Simonyan et al. \cite{simonyan2014very}} & \multirow{3}{*}{VGG19} & UFPR04 & 80.40 & 92.30 & 91.90 \\
& & UFPR05 & 88.80 & 95.10 & 95.90 \\
& & PUCPR & 81.50 & 93.80 & 94.60 \\
\hline
\multirow{3}{*}{He et al. \cite{he2016deep}} & \multirow{3}{*}{ResNet50} & UFPR04 & 93.70 & 94.80 & 93.30 \\
& & UFPR05 & 92.20 & 94.80 & 95.50 \\
& & PUCPR & 90.50 & 93.90 & 94.10 \\
\hline
\multirow{3}{*}{De et al. \cite{de2015pklot}} & \multirow{3}{*}{LPQ, LBP and its variants} & UFPR04 & 84.25 & 99.55 & 84.92 \\
& & UFPR05 & 87.74 & 85.76 & 98.90 \\
& & PUCPR & 99.58 & 87.15 & 82.78 \\
\hline
\multirow{3}{*}{De et al. \cite{de2015pklot}} & \multirow{3}{*}{Mean Ensemble} & UFPR04 & 88.40 & \textbf{99.64} & 88.33 \\
& & UFPR05 & 89.83 & 85.53 & 99.30 \\
& & PUCPR & 99.61 & 88.88 & 84.20 \\
\hline
\multirow{3}{*}{Nurullayev et al. \cite{nurullayev2019generalized}} & \multirow{3}{*}{CarNet} & UFPR04 & 98.30 & 95.60 & 97.60 \\
& & UFPR05 & 98.40 & 95.20 & 97.50 \\
& & PUCPR & 98.80 & 94.40 & 97.70 \\
\hline
\multirow{3}{*}{Amato et al. \cite{b10}} & \multirow{3}{*}{mAlexnet} & UFPR04 & 98.27 & 99.54 & 93.29 \\
& & UFPR05 & 92.72 & 93.69 & \textbf{99.49} \\
& & PUCPR & \textbf{99.90} & 98.03 & 96.00 \\
\hline
\multirow{3}{*}{Chollet et al. \cite{chollet}
} & \multirow{3}{*}{Xception} & UFPR04 & 94.00 & 94.60 & 93.40 \\
& & UFPR05 & 95.70 & 90.90 & 91.20 \\
& & PUCPR & 96.30 & 92.50 & 93.30 \\
\hline
\multirow{3}{*}{Szegedy et al. \cite{szegedy2016rethinking}
} & \multirow{3}{*}{Inception V3} & UFPR04 & 91.70 & 95.20 & 92.40 \\
& & UFPR05 & 94.30 & 92.90 & 93.70 \\
& & PUCPR & 90.80 & 91.10 & 94.20 \\
\hline
\multirow{3}{*}{Ratko Grbić, Brando Koch \cite{Koch}} & \multirow{3}{*}{APSD-OC} & UFPR04 & \textbf{99.98} & 95.47 & 99.19 \\
& & UFPR05 & 95.29 & \textbf{99.92} & 98.08 \\
& & PUCPR & 98.62 & \textbf{98.60} & \textbf{99.93} \\
\hline
\multirow{3}{*}{Proposed Model} & \multirow{3}{*}{ResNet-PinGTSVM} & UFPR04 & - & 98.40 & \textbf{99.70} \\
& & UFPR05 & \textbf{99.50} & - & 98.60 \\
& & PUCPR & 94.60 & 97.20 & - \\
\hline
\end{tabular}
\label{table:16}
\end{table}

\subsection{Dehazer Removal and Noise Reduction}

 Haze obscures image details by reducing contrast and introducing a grayish tone. Traditional dehazing methods focus on enhancing image clarity by removing these artifacts. However, robust machine learning models like pin-GTSVM can inherently handle noise and haze, reducing or eliminating the need for explicit dehazing steps.
 
AOD-NET \cite{AODNET}, i.e., All in One Dehazing Network, was used earlier for dehazing hazy images, after which they were fed as an input to a CNN to classify the images accordingly. In this paper, we eliminate the need for a ``Dehazer”, which is possible due to the following advantages:

\begin{itemize}
\item[i.]	\textbf{Robustness to noisy and outlier-prone data}: Using the pinball loss function, TWSVM is inherently designed to handle deviations from the target distribution, which often include noisy or outlier data points. This robustness reduces the reliance on preprocessing steps such as dehazing, typically aimed at mitigating the effects of noise and outliers in the data.

\item[ii.]	\textbf{Feature representation learning}: The feature extraction process, typically performed using pre-trained models like ResNet-50, enables extracting high-level features more robust to environmental factors such as haze. These features capture meaningful patterns and structures in the data, aiding in classification tasks without explicit dehazing.

\item[iii.]	\textbf{Focus on classification rather than image enhancement}: TWSVM focuses on optimizing the decision boundary between classes rather than directly enhancing the visual quality of input images. While dehazing algorithms like AOD-Net aim to improve image clarity by removing haze, TWSVM addresses the classification task without altering the input images. This streamlined approach eliminates the need for separate dehazing steps in the workflow.

\item[iv.]	\textbf{Generalization across diverse datasets:} TWSVM, particularly when combined with feature extraction from pre-trained models, exhibits strong generalization capabilities across diverse datasets and environmental conditions. Learning discriminative features from various inputs makes the model less sensitive to specific environmental factors like haze, reducing the need for specialized preprocessing techniques like dehazing.

\item[v.]  \textbf{Implicit Dehazing:} When training the model, the inherent robustness of the pinball loss function helps the SVM focus on the underlying patterns rather than the noisy variations. This results in the hyperplanes that generalize well to clean data, effectively acting as an implicit dehazing mechanism. The model learns to ignore the noise (haze) and focus on the true data distribution.

\item[vi.]   \textbf{A significant jump in Accuracy:} The replacement of a dehazer network with the Pin-GTSVM classifier and a Pre-trained Feature Extractor shows a major difference in the classification of Hazy Images. The previously obtained highest accuracy, Model 3 ($\lambda$=0.4290) in \ref{table:6} was 88.63(\%), and we obtained better high accuracies, as shown in the result analysis section.
\end{itemize}
Overall, the robustness, feature representation learning capabilities, and focus on classification tasks inherent in the TWSVM approach contribute to minimizing the need for a separate dehazing step in the workflow. This not only simplifies the overall methodology but also enhances the efficiency and effectiveness of the classification process, particularly in scenarios where environmental factors like haze may pose challenges to traditional image processing techniques.

\section{Conclusion}
This study introduces a parking space occupancy detection model capable of operating effectively in hazy environments and other challenging atmospheric conditions. The system is divided into two primary components: A Feature Extractor using pre-trained models like ResNet-50, GoogleNet, and AlexNet and classification using Pin-GTSVM. The system's performance is evaluated on a diverse set of images, including both hazy and non-hazy conditions. Compared to other state-of-the-art Algorithms, it shows supremacy in all the benchmarks, such as Accuracy, Precision, Recall, F-1 scores, and Specificity values. We conclude that this novel hybrid pre-trained CNN and pin-GTSVM Model removes the need for a separate Dehazer and provides a one-of-a-kind parking slot classifier. This work can be further enhanced by incorporating quantization techniques, which can significantly reduce the runtime of the proposed system. Additionally, exploring advanced optimization methods or hardware acceleration could further improve efficiency and scalability, making the system more practical for real-world applications. Also, we can further investigate the proposed model's noise resilience across various real-world applications, including medicine, healthcare, and cybersecurity.

\section*{Declarations}
\begin{itemize}
\item Funding: There is no funding associated to this study.
\item Conflict of interest/Competing interests: There is no conflict of interest. 
\item Ethics approval and consent to participate: Not applicable.
\item Consent for publication: All authors consent to the publication of this paper.
\item Data availability: The datasets generated during and/or analyzed during the current study are available from the corresponding author upon reasonable request.
\item Materials availability: Materials used in this study are available from the corresponding author upon reasonable request.
\item Code availability: The code used in this study is available from the corresponding author on reasonable request.
\item Author contribution: All authors have contributed equally to the work.
\end{itemize}
\section*{Abbreviations}
Table \ref{tab:abr} lists all the abbreviations used in the paper.

\begin{table}[h!]
\caption{Abbreviations used in the paper}
    \label{tab:abr}
    \centering
    \scriptsize
    \begin{tabular}{|lp{5.5cm}|lp{5.5cm}|}
    \hline
     CNN &Convolution Neural Network &Pin-GTSVM &Pinball Generalized Twin Support Vector Machines   \\
TWSVM & Twin Support Vector Machine & RBF &Radial Basis Function\\
RF & Random Forest & PUCPR & Pontifical Catholic University of Paran´\\
AOD-Net & All-in-One Dehazing Network& 
SVM & Support Vector Machine \\
LRN & Local Response Normalization & UFPR & Federal University of Paraná \\
ResNet & Residual Network& ASM& Atmospheric Scattering Model\\
ReLU & Rectified Linear Unit & tanh &hyperbolic tangent \\
FCL & Fully Connected Layer & CL & Convolution Layer \\
MLELM & Multilayer ELM & KNN & K-Nearest  \\
GNB & Gaussian Naive-Bayes & XGBoost & Extreme Gradient Boosting \\
LR & Linear Regression &  VGG & Visual Geometry Group \\
\hline
\end{tabular}    
\end{table}
\noindent
\bibliography{Reference_1}

\end{document}